\documentclass[sigconf, nonacm]{acmart}

\usepackage{hyperref}       
\usepackage{url}            
\usepackage{booktabs}       
\usepackage{amsfonts}       
\usepackage{nicefrac}       

\usepackage{rotating}
\usepackage{multicol} 

\usepackage[commentsnumbered,linesnumbered,ruled]{algorithm2e}

\usepackage{caption}
\usepackage{subcaption}

\usepackage{enumitem}
\usepackage{amsmath}
\usepackage{amsthm}
\usepackage{bigints}
\usepackage{dsfont}
\usepackage{bbm}
\usepackage{textcomp}
\usepackage{stmaryrd}

\usepackage{mathtools}
\usepackage{graphicx}
\usepackage{combinedgraphics}
\graphicspath{{graphs/}{graphs_extra/}}

\usepackage{multirow}
\usepackage{tabularx}
\usepackage{arydshln}

\usepackage{array,etoolbox}
\preto\tabular{\setcounter{magicrownumbers}{0}}
\newcounter{magicrownumbers}

\newtheorem{thm}{Theorem}
\newtheorem{lem}{Lemma}
\newtheorem{Athm}{Theorem}
\newtheorem{Alem}{Lemma}
\newtheorem{col}{Corollary}

\newtheorem{prop}{Proposition}
\newtheorem{definition}{Definition}

\begin{document}
	
	\title{Breaking the curse of dimensionality  with Isolation Kernel}

	\author{Kai Ming Ting}
	\affiliation{%
		\institution{ Nanjing University}
	}
	\email{tingkm@nju.edu.cn}
	
	\author{Takashi Washio}
	\affiliation{%
		\institution{ Osaka University}
	}
	\email{washio@ar.sanken.osaka-u.ac.jp}
	
	\author{Ye Zhu}
	\affiliation{%
		\institution{Deakin University}
	}
	\email{ye.zhu@ieee.org}

	\author{Yang Xu}
	\affiliation{%
		\institution{ Nanjing University}
	}
	\email{xuyang@lamda.nju.edu.cn}

	\begin{abstract}
		The curse of dimensionality has been studied in different aspects.
		However, breaking the curse has been elusive. We show for the first time that it is possible to break the curse using the recently introduced Isolation Kernel. We show that only Isolation Kernel performs consistently well in indexed search, spectral \& density peaks clustering, SVM classification and t-SNE visualization in both low and high dimensions, compared with distance, Gaussian and linear kernels. This is also supported by our theoretical analyses that Isolation Kernel is the only kernel that has the provable ability to break the curse, compared with existing metric-based Lipschitz continuous kernels. 
	\end{abstract}

	\maketitle
	
	\section{Introduction}
	
	In the last few decades, a significant amount of research has been invested in studying the curse of dimensionality of a distance measure. We have now a good understanding of the curse in terms of (i) concentration of measure \citep{Talagrand} and its consequence of being unable to find the nearest neighbor \citep{Beyer:1999}, and (ii) hubness \citep{HubsInSpace-JMLR-2010}.
	
	Despite this advancement, breaking the curse is still elusive. The central question remains open:   \textit{is it possible to find the single nearest neighbor of a query in high dimensions?}
	
	The answer to this question is negative, based on the current analyses on distance \citep{Beyer:1999,Density-based-Indexing-KDD1999,Pestov}. Rather than finding the single nearest neighbor of a query, the issue has been diverted to finding all data points belonging to the same cluster as the query, in a distribution with well-separated clusters \citep{Beyer:1999,Density-based-Indexing-KDD1999}. These nearest neighbors are `meaningful' for the purpose of separating different clusters in a dataset. But the (single) nearest neighbor, in the true sense of the term, cannot be obtained using a distance measure, even in the special distribution of well-separated clusters described above \citep{Beyer:1999}.
	
	
	Unlike many existing measures, a recently introduced Isolation Kernel \citep{ting2018IsolationKernel,IsolationKernel-AAAI2019} is derived directly from a finite dataset without learning and has no closed-form expression.
	We show here that Isolation Kernel is not affected by the concentration effect, and it works well in high dimensions; and it does not need to use low intrinsic dimensions to explain why it works well in high dimensions.

	Our theoretical analysis focuses on \textit{the ability of a feature map produced by kernel evaluation using a finite dataset to distinguish two distinct points  in high dimensions} (\textbf{distinguishability} or \textbf{indistinguishability} hereafter). A kernel is said to have indistinguishability if its feature map is unable to distinguish two distinct points. This is manifested as being unable to find the single nearest neighbor of a query because there are many nearest neighbors which cannot be distinguished from each other, especially in high dimensions. 
	
	Our contributions are:
	\vspace{-1mm}
	\renewcommand{\labelenumi}{\Roman{enumi}}
	\begin{enumerate}[itemsep=1ex, leftmargin=5mm]
		\item  Revealing that Isolation Kernel (IK) is the only measure that is provably free from the curse of dimensionality, 
		compared with existing metric-based Lipschitz continuous kernels.
		
		\item Determining three key factors in IK that lead to breaking the curse of dimensionality.
		(a) Space partitioning using an isolation mechanism is the key ingredient in measuring similarity between two points in IK. 
		(b) The probability of a point falling into a partition
		is independent of data distribution, the distance measure used to create the partitions and the data dimensions.
		(c) The unique feature map of IK has its dimensionality linked to a concatenation of these partitionings. 
		
		\item Proving that IK's unique feature map produced by a finite dataset has distinguishability,
		independent of the number of data dimensions and data distribution. 
		Our theorem suggests that increasing the number of partitionings (i.e., the dimensionality of the feature map) leads to increased distinguishability.
		
		\item Verifying that IK 
		breaks the curse of dimensionality in two sets of experiments. First, IK finds the single nearest neighbor in high dimensions; while many existing measures fail under the same condition. Second, IK enables consistently effective indexing, spectral \& density peaks clustering, SVM classification and t-SNE visualization in both low and high dimensions.
		
		
		\item Showing that Euclidean distance, Gaussian and linear kernels have a mixed bag of poor and good results in high dimensions, reflecting the current understanding of existing measures.
	\end{enumerate}

	\section{Related work}
	
	\subsection{The curse of dimensionality}
	\label{sec_curse}
	One of the early investigations of the curse of dimensionality is in terms of concentration of measure \cite{Mil_man_1972,Talagrand}. \citet{Talagrand} summarises the intuition behind the concentration of measure
	as follows: ``a random variable that depends (in a `smooth' way) on the influence of many independent
	random variables (but not too much on any of them) is essentially constant.''
	Translating this statement in the context of a high dimensional space, a distance measure which depends on many independent dimensions (but not too much on any of them) is essentially constant.
	
	\citet{Beyer:1999} analyze the conditions under which nearest neighbor is `meaningful' when a distance measure is employed
	in high dimensions.
	A brief summary of the conditions in high dimensions is given as follows:
	\renewcommand{\labelenumi}{\alph{enumi})}
	\begin{enumerate}
		\item Finding the (single) nearest neighbor of a query is not meaningful because every point from the same cluster in the database has approximately the same distance to the query. In other words, the variance of distance distribution within a cluster approaches zero as the number of dimensions increases to infinity.
		\item Despite this fact, finding the exact match is still meaningful.
		\item The task of retrieving all points belonging to the same cluster as the query is meaningful because although variance of distances of all points within the cluster is zero, the distances of this cluster to the query are different from those of another cluster---making a distinction between clusters possible.
		\item The task of retrieving the nearest neighbor of a query, which does not belong to any of the clusters in the database, is not meaningful for the same reason given in (a).
	\end{enumerate}
	\citet{Beyer:1999} reveal that finding (non-exact match) nearest neighbors is `meaningful' only if the dataset has clusters (or structure), where the nearest neighbors refer to any/all points in a cluster rather than a specific point which is the nearest among all points in the dataset. In short, a distance measure cannot produce the `nearest neighbor' in high dimensions, in the true sense of the term. Hereafter the term `nearest neighbor' is referred to the single nearest neighbor in a dataset, not many nearest neighbors in a cluster.
	
	\citet{Beyer:1999} describe the notion of instability of a distance measure as follows: ``{\em A nearest neighbor query is unstable for a given $\epsilon$ if the distance from the query point to most data points is less than (1 + $\epsilon$) times the distance from the query point to its nearest neighbor.}''

	
	Current analyses on the curse of dimensionality focus on distance measures \citep{Beyer:1999,NN-HighDim-VLDB-2000,Aggarwal-2001,HubsInSpace-JMLR-2010}. While random projection preserves structure in the data, it still suffers the concentration effect \citep{Ata-Kaban-2011}. 
	Current methods dealing with high dimensions are: dimension reduction \citep{PCA-Book2002,Hinton2008Visualizing}; suggestions to use non-metric distances, e.g., fractional distances \citep{Aggarwal-2001}; and angle-based measures, e.g., \citep{Angle-BasedAD-KDD2008}. None has been shown to break the curse.
	
	Yet, the influence of the concentration effect in practice is less severe than suggested mathematically \citep{Concentration-Fractionaldistances}. For example,  k-nearest neighbor (kNN) classifier still yields reasonable high accuracy in high dimensions \citep{Aryal2017}; while the concentration effect would have rendered kNN producing random prediction. One possible explanation of this phenomenon is that the high dimensional datasets employed have low intrinsic dimensions (e.g., \citep{Concentration-Fractionaldistances,LID-NIPS2011}.) 
	
	
	
	
	In a nutshell, existing works center around reducing the effect of concentration of measures,
	knowing that the  nearest neighbor could not be obtained in high dimensions. 
	A saving grace is that many high dimensional datasets used in practice have low intrinsic dimensions---leading to a significantly reduced concentration effect.
	
	Our work shows that the concentration effect can be eliminated to get the  nearest neighbor in high dimensions using a recently introduced measure called Isolation Kernel \citep{ting2018IsolationKernel,IsolationKernel-AAAI2019}, without the need to rely on low intrinsic dimensions. While three existing measures have a mixed bag of poor and good results in high dimensions, Isolation Kernel defined by a finite dataset yields consistently good results in both low and high dimensions.
	
	\subsection{Isolation Kernel}
	
	Unlike commonly used distance or kernels, Isolation Kernel \citep{ting2018IsolationKernel,IsolationKernel-AAAI2019} has no closed-form expression, and it is derived directly from a given dataset without learning; and the similarity between two points is computed based on the partitions created in data space.
	
	The key requirement of Isolation Kernel (IK) is a space partitioning mechanism which isolates a point from the rest of the points in a sample set. This can be achieved in several ways, e.g., isolation forest \citep{liu2008isolation,ting2018IsolationKernel}, Voronoi Diagram \citep{IsolationKernel-AAAI2019} and isolating hyperspheres \citep{IDK-KDD2020}. 
	Isolation Kernel \citep{ting2018IsolationKernel,IsolationKernel-AAAI2019} is defined as follows.
	
	Let $\mathbb{H}_\psi(D)$ denote the set of all admissible partitionings $H$ derived from a given finite dataset $D$, where each $H$ is derived from $\mathcal{D} \subset D$; and each point in $\mathcal{D}$ has the equal probability of being selected from $D$; and $|\mathcal{D}|=\psi$.
	Each partition $\theta[{\bf z}] \in H$ isolates a point ${\bf z} \in \mathcal{D}$ from the rest of the points in $\mathcal{D}$. The union of all  partitions of each partitioning $H$ covers the entire space.

	\begin{definition}   
		Isolation Kernel of any two points ${\bf x},{\bf y} \in \mathbb{R}^d$ is defined to be
		the expectation taken over the probability distribution on all partitionings $H \in \mathds{H}_\psi(D)$ that both ${\bf x}$ and ${\bf y}$  fall into the same isolating partition $\theta[{\bf z}] \in H$, where ${\bf z} \in \mathcal{D} \subset D$, $\psi=|\mathcal{D}|$:
		\begin{eqnarray}
			K_\psi({\bf x},{\bf y}\ |\ D)  =  {\mathbb E}_{\mathds{H}_\psi(D)} [\mathds{1}({\bf x},{\bf y} \in \theta[{\bf z}]\ | \ \theta[{\bf z}] \in H)],
			\label{eqn_kernel}
		\end{eqnarray}
	\end{definition}
	\noindent
	where $\mathds{1}(\cdot)$ is an indicator function.
	
	In practice, Isolation Kernel $K_\psi$ is constructed using a finite number of partitionings $H_i, i=1,\dots,t$, where each $H_i$ is created using randomly subsampled $\mathcal{D}_i \subset D$; and $\theta$ is a shorthand for $\theta[{\bf z}]$:
	\begin{eqnarray}
		K_\psi({\bf x},{\bf y}\ |\ D)   & \simeq &  \frac{1}{t} \sum_{i=1}^t   \mathds{1}({\bf x},{\bf y} \in \theta\ | \ \theta \in H_i) \nonumber\\
		& = &   \frac{1}{t} \sum_{i=1}^t \sum_{\theta \in H_i}   \mathds{1}({\bf x}\in \theta)\mathds{1}({\bf y}\in \theta). 
		\label{Eqn_IK}
	\end{eqnarray}
	This gives a good approximation of $K_\psi({\bf x},{\bf y}\ |\ D)$ when $|D|$ and $t$ are sufficiently large to ensure that the ensemble is obtained from a sufficient number of mutually independent $\mathcal{D}_i, i=1,\dots,t$.
	
	Given $H_i$, let $\Phi_i({\bf x})$ be a $\psi$-dimensional binary column (one-hot) vector representing all $\theta_j \in H_i$, $j=1,\dots,\psi$; where ${\bf x}$ must fall into only one of the $\psi$ partitions.
	
	The $j$-component of the vector is:
	$\Phi_{ij}({\bf x})=\mathds{1}({\bf x}\in \theta_j\ |\ \theta_j\in H_i)$. Given $t$ partitionings, $\Phi({\bf x})$ is the concatenation of $\Phi_1({\bf x}),\dots,\Phi_t({\bf x})$.
	
	\begin{prop}
		\label{prop:featureMap}
		\textbf{Feature map of Isolation Kernel.}
		The feature mapping $\Phi: {\bf x}\rightarrow \mathbb \{0,1\}^{t\times \psi}$ of $K_\psi$, for point ${\bf x} \in \mathbb{R}^d$, is a vector that represents the partitions in all the partitionings $H_i\in \mathds{H}_\psi(D)$, $i=1,\dots,t$ that contain ${\bf x}$; where ${\bf x}$ falls into only one of the $\psi$ partitions in each partitioning $H_i$. Then, $\parallel {\Phi}({\bf x}) \parallel\ = \sqrt{t}$ holds, and Isolation Kernel always computes its similarity using its feature map as:
		\[
		K_\psi({\bf x},{\bf y}\ |\ D)   \simeq  \frac{1}{t} \left< {\Phi}({\bf x}|D), {\Phi}({\bf y}|D) \right>.
		\]
		
	\end{prop}

	Let $\mathbbmtt{1}$ be a shorthand of $\Phi_i({\bf x})$ such that $\Phi_{ij}({\bf x})=1$ and $\Phi_{ik}({\bf x})=0, \forall k \neq j$ of any $j \in [1,\psi]$. 
	
	The feature map
	${\Phi}({\bf x})$ is sparse because it has exactly $t$ out of $t \psi$ elements having 1 and the rest are zeros. This gives $\parallel {\Phi}({\bf x}) \parallel\ = \sqrt{t}$. 
	Note that  while every point has $\parallel {\Phi}({\bf x}) \parallel\ = \sqrt{t}$ and $\Phi_i({\bf x}) = \mathbbmtt{1}$ for all $i \in [1,t]$, they are not all the same point.
	Note that Isolation Kernel is a positive definite kernel with probability 1 in the limit of $t \rightarrow \infty$ because its Gram matrix  is full rank as $\Phi(x)$ for all points $x \in D$ are mutually independent (see \cite{IDK-KDD2020} for details.) In other words, the feature map coincides with a reproducing kernel Hilbert space (RKHS) associated with the Isolation Kernel.
	
	This finite-dimensional feature map has been shown to be the key in leading to fast runtime in kernel-based anomaly detectors \citep{IDK-KDD2020} and enabling the use of fast linear SVM \citep{IsolationSetKernel,IK-XFactor-2019}.
	
	IK has been shown to be better than Gaussian and Laplacian kernels in SVM classification \citep{ting2018IsolationKernel}, better than Euclidean distance in density-based clustering \citep{IsolationKernel-AAAI2019}, and better than Gaussian kernel in kernel-based anomaly detection \citep{IDK-KDD2020}.
	
	IK's superiority has been attributed to its data dependent similarity, given in the following lemma.
	
	Let $D \subset \mathcal{X}_D \subseteq {\mathbb R}^d$ be a dataset sampled from an unknown distribution $\mathcal{P}_D$ where $\mathcal{X}_D$ is the support of $\mathcal{P}_D$ 
	on ${\mathbb R}^d$; and
	let $\rho_D({\bf x})$ denote the density of $\mathcal{P}_D$ at point ${\bf x} \in \mathcal{X}_D$, and $\ell_p$ be the $\ell_p$-norm. 
	The unique characteristic of Isolation Kernel \citep{ting2018IsolationKernel,IsolationKernel-AAAI2019} is:
	\begin{lem}\label{lem_characteristic} 
		Isolation Kernel $K_\psi$ has the data dependent characteristic:
		\vspace{-2mm}
		\begin{equation}
			K_\psi( {\bf x},{\bf y}\ |\ D) > K_\psi( {\bf x}',{\bf y}'\ |\ D)\nonumber
		\end{equation}
		for $\ell_p({\bf x}-{\bf y})\ =\-\ell_p({\bf x}'-{\bf y}')$,
		$\forall {\bf x},{\bf y} \in \mathcal{X}_\mathsf{S}$ and $\forall {\bf x}',{\bf y}' \in  \mathcal{X}_\mathsf{T}$ 
		subject to $\forall {\bf z}\in \mathcal{X}_\mathsf{S}, {\bf z}'\in \mathcal{X}_\mathsf{T}, \ \rho_D({\bf z})<\rho_D({\bf z}')$.
	\end{lem}
	
	As a result, $K_\psi( {\bf x},{\bf y}\ |\ D)$ is not translation invariant, unlike the data independent Gaussian and Laplacian kernels.
	
	We show in Section \ref{sec_distinguishability} that this feature map is instrumental in making IK the only measure that breaks the curse,  compared with existing metric-based Lipschitz continuous kernels.

	\section{Analyzing the curse of dimensionality in terms of (in)distinguishability}
	\label{sec_break-curse}
	In this section, we theoretically show that IK does not suffer from the concentration effect in any high dimensional data space;
	while metric-based Lipschitz continuous kernels, including  Gaussian and Laplacian kernels, do. 
	
	We introduce the notion of indistinguishability of two distinct points in a feature space $\mathcal{H}$ produced by kernel evaluation
	using a finite dataset
	to analyze the curse of dimensionality. Let $\Omega \subset \mathbb{R}^d$ be a data space, $D \subset \Omega$ be a finite dataset, and $\mathcal{P}_D$ be the probability distribution that generates $D$. The support of $\mathcal{P}_D$, denoted as $\mathcal{X}_D$, is an example of $\Omega$. 
	
	Given two 
	distinct points ${\bf x}_a,{\bf x}_b \in \Omega$ and their feature vectors 
	$\phi({\bf x}_a), \phi({\bf x}_b) \in \mathcal{H}_\kappa$ produced by a kernel $\kappa$ and a finite dataset $D = \{{\bf x}_i \in \mathbb{R}^d \; | \; i=1, \dots, n\}$ as $\phi({\bf x}_a) = [\kappa({\bf x}_a,{\bf x}_1),\dots,\kappa({\bf x}_a,{\bf x}_n)]^\top$ and 
	$\phi({\bf x}_b) = [\kappa({\bf x}_b,{\bf x}_1),\dots,\kappa({\bf x}_b,{\bf x}_n)]^\top$. 
	
	\begin{definition}
		$\mathcal{H}_\kappa$ has \textbf{indistinguishability} if 
		there exists some $\delta_L \in (0,1)$ such that 
		the probability of $\phi({\bf x}_a) = \phi({\bf x}_b)$ is lower bounded by $\delta_L$ as follows: 
		\begin{equation}
			P(\phi({\bf x}_a) = \phi({\bf x}_b)) \geq \delta_L. \label{indist}
		\end{equation}
	\end{definition}
	
	On the other hand, given two 
	feature vectors 
	$\Phi({\bf x}_a), \Phi({\bf x}_b) \in \mathcal{H}_K$ produced by a kernel $K$ and a finite dataset $D$. 
	
	\begin{definition}
		$\mathcal{H}_K$  has \textbf{distinguishability} if there exists some $\delta_U \in (0,1)$ such that $P(\Phi({\bf x}_a) = \Phi({\bf x}_b))$ is upper bounded by $\delta_U$ as follows: 
		\begin{equation}
			P(\Phi({\bf x}_a) = \Phi({\bf x}_b)) \leq \delta_U. \label{eqn_distinguishability}
		\end{equation}
	\end{definition}
	Note that these definitions of indistinguishability and distinguishability are rather weak. In practice, $\delta_L$ should be close to 1 for strong indistinguishability, and $\delta_U$ should be far less than 1 for strong distinguishability.
	
	In the following, we first show in Section \ref{sec_indistinguishability} that a feature space $\mathcal{H}_\kappa$ produced by any metric-based Lipschitz 
	continuous kernel $\kappa$ and a finite dataset $D$ has indistinguishability when the number of dimensions $d$ of the data space is very large. 
	Then, in Section \ref{sec_distinguishability}, we show that a feature space $\mathcal{H}_K$ procudd by Isolation Kernel $K$ implemented using the Voronoi diagram \citep{IsolationKernel-AAAI2019}
	has distinguishability for any $d$.
	
	\subsection{$\mathcal{H}_\kappa$ of a metric-based Lipschitz 
		continuous kernel $\kappa$ has indistinguishability}
	\label{sec_indistinguishability}
	
	Previous studies have analyzed the concentration
	effect in high dimensions~\citep{Pestov, Gromov, Ledoux}. Let a triplet 
	$(\Omega, m, F)$ be a probability metric space where $m: \Omega \times \Omega \mapsto \mathbb{R}$ 
	is a metric, and 
	$F: 2^\Omega \mapsto \mathbb{R}$ is a probability measure.  $\ell_p$-norm is an 
	example of $m$. $\Omega$ is the support of $F$, and thus $F$ is assumed to have non-negligible probabilities over the entire $\Omega$, i.e., points drawn from $F$ are widely distributed over the entire data space. 
	
	Further, let $f: \Omega \mapsto \mathbb{R}$ be a Lipschitz continuous function, i.e., $|f({\bf x}_a) - f({\bf x}_b)| \leq m({\bf x}_a,{\bf x}_b), \forall {\bf x}_a,{\bf x}_b \in \Omega$; 
	and $M$ be a median of $f$ defined as:
	\begin{eqnarray}
		F(\{{\bf x} \in \Omega \; | \; f({\bf x}) \leq M\}) = F(\{{\bf x} \in \Omega \; | \; f({\bf x}) \geq M\}).\nonumber
	\end{eqnarray}
	Then, the following proposition holds~\citep{Pestov, Gromov}.
	\begin{prop}\label{prop1}
		$F(\{{\bf x} \in \Omega | f({\bf x}) \in [M-\epsilon,M+\epsilon]\}) \geq 1-2\alpha(\epsilon)$ holds for any $\epsilon>0$, where 
		$\alpha(\epsilon) = C_1 e^{-C_2 \epsilon^2 d}$ with two constants $0 < C_1 \leq \frac{1}{2}$ and $C_2 > 0$.
	\end{prop}
	
	For our analysis on the indistinguishability of a feature space $\mathcal{H}_\kappa$ associated with any metric-based Lipschitz continuous kernel, we reformulate this proposition 
	by replacing $f({\bf x})$ and $M$ with $m({\bf x},{\bf y})$ and $M({\bf y})$, respectively, for 
	a given ${\bf y} \in \Omega$. That is:
	\begin{eqnarray}
		&&\hspace*{-5mm} |m({\bf x}_a,{\bf y}) - m({\bf x}_b,{\bf y})| \leq m({\bf x}_a,{\bf x}_b), \forall {\bf x}_a,{\bf x}_b \in \Omega, \mbox{ and}\nonumber\\
		&&\hspace*{-5mm}F(\{{\bf x} \in \Omega \; | \; m({\bf x},{\bf y}) \leq M({\bf y})\}) = F(\{{\bf x} \in \Omega \; | \; m({\bf x},{\bf y}) \geq M({\bf y})\}).\nonumber
	\end{eqnarray}
	The inequality in the first formulae always holds because of the triangular inequality of $m$. 
	Then, we obtain the following corollary: 
	\begin{col}\label{col1}
		Under the same condition in Proposition~\ref{prop1}, the inequality $F(A_\epsilon({\bf y})) \geq 1- 2\alpha(\epsilon)$ holds for any ${\bf y} \in \Omega$ 
		and any $\epsilon > 0$, where 
		$A_\epsilon({\bf y}) = \{{\bf x} \in \Omega \; | \; m({\bf x},{\bf y}) \in [M({\bf y})-\epsilon, M({\bf y})+\epsilon]\}$. 
	\end{col}
	
	This corollary represents the  concentration effect where almost all 
	probabilities of the data are concentrated in the limited area $A_\epsilon({\bf y})$. 
	It provides an important fact on the indistinguishability of two distinct points using 
	a class of metric-based Lipschitz continuous kernels such as Gaussian kernel:
	\begin{equation}
		\kappa({\bf x},{\bf y}) = f_\kappa(m({\bf x},{\bf y})),\nonumber
	\end{equation} 
	where $f_\kappa$ is Lipschitz continuous and monotonically decreasing for $m({\bf x},{\bf y})$.
	\begin{lem}\label{lem1}
		Given two distinct points ${\bf x}_a$, ${\bf x}_b$ i.i.d. drawn from $F$ on $\Omega$ 
		and a data set $D = \{{\bf x}_i \in \mathbb{R}^d \; | \; i=1, \dots, n\} \sim G^n$ 
		where every ${\bf x}_i$ is i.i.d drawn from any probability distribution $G$ on 
		$\Omega$, let the feature vectors of ${\bf x}_a$ and ${\bf x}_b$ be 
		$\phi({\bf x}_a) = [\kappa({\bf x}_a,{\bf x}_1),\dots,\kappa({\bf x}_a,{\bf x}_n)]^\top$ and 
		$\phi({\bf x}_b) = [\kappa({\bf x}_b,{\bf x}_1),\dots,\kappa({\bf x}_b,{\bf x}_n)]^\top$ in a feature space $\mathcal{H}_\kappa$.  The following holds:
		\begin{equation}
			P\left(\ell_p(\phi({\bf x}_a)-\phi({\bf x}_b)) \leq 2L\epsilon n^{1/p} \right) \geq (1-2\alpha(\epsilon))^{2n}, \nonumber
		\end{equation}
		for any $\epsilon > 0$, where $\ell_p$ is an $\ell_p$-norm on $\mathcal{H}_\kappa$, and 
		$L$ is the Lipschitz constant of $f_\kappa$.
	\end{lem}
	
	\begin{thm}\label{thm1}
		Under the same condition as Lemma~\ref{lem1}, the feature space $\mathcal{H}_\kappa$ 
		has indistinguishability in the limit of $d \rightarrow \infty$.
		
	\end{thm}
	
	This result implies that the feature space $\mathcal{H}_\kappa$ produced by any finite dataset and any metric-based 
	Lipschitz continuous kernels such as Gaussian and Laplacian kernels have indistinguishability 
	when the data space has a high number of dimensions $d$. In other words, these kernels suffer 
	from the concentration effect.
	
	\subsection{$\mathcal{H}_K$ of Isolation Kernel $K$ has distinguishability}
	\label{sec_distinguishability}
	We show that the feature map of Isolation Kernel implemented using Voronoi diagram \citep{IsolationKernel-AAAI2019}, which is not a metric-based 
	Lipschitz continuous kernel, has distinguishability.
	
	We derive Lemma \ref{lem2} from the
	fact given in \cite{Fukunaga}, before providing our theorem.
	\begin{lem}\label{lem2}
		Given ${\bf x} \in \Omega$ and a data set 
		$\mathcal D = \{ {\bf z}_1, \dots, {\bf z}_\psi \} \sim G^\psi$ where 
		every ${\bf z}_j$ is i.i.d. drawn from any probability distribution $G$ on 
		$\Omega$ and forms its Voronoi partition $\theta_j \subset \Omega$, the probability that ${\bf z}_j$ is the nearest neighbor 
		of ${\bf x}$ in $\mathcal D$ is given as: $P({\bf x} \in \theta_j) = 1/\psi$, for every $j=1,\dots,\psi$.  
	\end{lem}
	This lemma points to a simple but nontrivial result that
	$P({\bf x} \in \theta_j)$ 
	is independent of $G$, $m({\bf x},{\bf y})$ and data dimension $d$. 
	
	\begin{thm}\label{thm2}
		Given two distinct points ${\bf x}_a$, ${\bf x}_b$ i.i.d. drawn from $F$ on $\Omega$ 
		and ${\mathcal D}_i = \{ {\bf z}_1, \dots, {\bf z}_\psi \} \sim G^\psi$ defined in 
		Lemma~\ref{lem2}, let the feature vectors of ${\bf x}_a$ and ${\bf x}_b$ be $\Phi({\bf x}_a)$ and $\Phi({\bf x}_b)$
		in a feature space $\mathcal{H}_K$ associated with Isolation Kernel $K$, implemented using Voronoi diagram, as given in Proposition \ref{prop:featureMap}. 
		Then the following holds: 
		\[
		P(\Phi({\bf x}_a)=\Phi({\bf x}_b)) \leq \cfrac{1}{\psi^t},
		\]
		and $\mathcal{H}_K$  always has strong distinguishability for large $\psi$ and $t$.
	\end{thm}
	
	Proofs of Lemmas \ref{lem1} \& \ref{lem2} and Theorems \ref{thm1} \& \ref{thm2} are given in Appendix \ref{sec_A_Proofs}.

	In summary, \textbf{the IK  implemented using Voronoi diagram is free from the curse of dimensionality for any probability distribution}. 
	
	\subsection{The roles of  $\psi$ \& $t$ and the time complexity of Isolation Kernel} 
	The sample size (or the number of partitions in one partitioning) $\psi$ in IK has a function similar to the bandwidth parameter in Gaussian kernel, i.e., the higher $\psi$ is, the sharper the IK distribution \citep{ting2018IsolationKernel} (the small the bandwidth, the sharper the Gaussian kernel distribution.) In other words, $\psi$ needs to be tuned for a specific dataset to produce a good task-specific performance.
	
	The number of partitionings $t$ has three roles. First, the higher  $t$ is, the better Eq \ref{Eqn_IK} is in estimating the expectation described in Eq \ref{eqn_kernel}. High $t$ also leads to low variance of the estimation. Second, with its feature map, $t$ can be viewed as IK's  user-definable effective number of dimensions in Hilbert space. Third, Theorem~\ref{thm2} reveals that  increasing $t$ leads to increased distinguishability. The first two roles were uncovered in the previous studies \citep{ting2018IsolationKernel,IsolationKernel-AAAI2019,IDK-KDD2020}. We reveal the last role here in Theorem \ref{thm2}. These effects can be seen in the experiment reported in Appendix \ref{sec_partitiongs}.
	
	In summary, \textbf{IK's feature map, which has its dimensionality linked to distinguishability, is unique among existing measures}. In contrast, Gaussian kernel (and many existing metric-based kernels) have a feature map with intractable dimensionality; yet, they have indistinguishability.
	
	\textbf{Time complexity}:
	IK, implemented using the Voronoi diagram, has time complexity $O(ndt\psi)$ to convert $n$  points in data space to $n$ vectors in its feature space. Note that each $\mathcal{D}$ implemented using one-nearest-neighbor produces a Voronoi diagram implicitly, no additional computation is required to build the Voronoi diagram explicitly (see \citet{IsolationKernel-AAAI2019} for details.) In addition, as this implementation is amenable to acceleration using parallel computing, it can be reduced by a factor of $w$ parallelizations to $O(\frac{n}{w}dt\psi)$. 
	The readily available finite-dimensional feature map of IK enables a fast linear SVM to be used. In contrast, Gaussian kernel (GK) must use a slow nonlinear SVM. Even adding the IK's feature mapping time, SVM with IK still runs up to two orders of magnitude faster than SVM with GK. See Section \ref{sec_traditional_tasks}  for details. 
	
	\section{Empirical verification}
	\label{sec_experiments}
	We verify that IK breaks the curse of dimensionality by conducting two experiments.
	In Section~\ref{sec_instability}, we verify the theoretical results in terms of instability of a measure, as used in a previous study of the concentration effect \citep{Beyer:1999}. This examines the (in)distinguishability of linear, Gaussian and Isolation kernels.
	In the second experiment in Section \ref{sec_traditional_tasks}, we explore the impact of IK having distinguishability in three tasks:  ball tree indexing \citep{omohundro1989five}, SVM classification \citep{CC01a} and t-SNE visualization \citep{Hinton2008Visualizing}.
	

	\begin{figure*}[t]
		\centering
		\begin{subfigure}[b]{0.245\textwidth}
			\centering
			\includegraphics[width=1.5in]{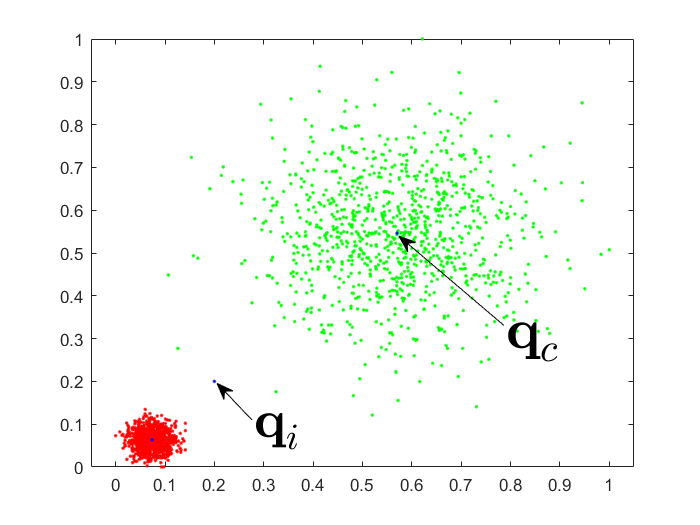}
			\caption{Dataset used}
		\end{subfigure}
		\begin{subfigure}[b]{0.245\textwidth}
			\centering
			\includegraphics[width=1.6in]{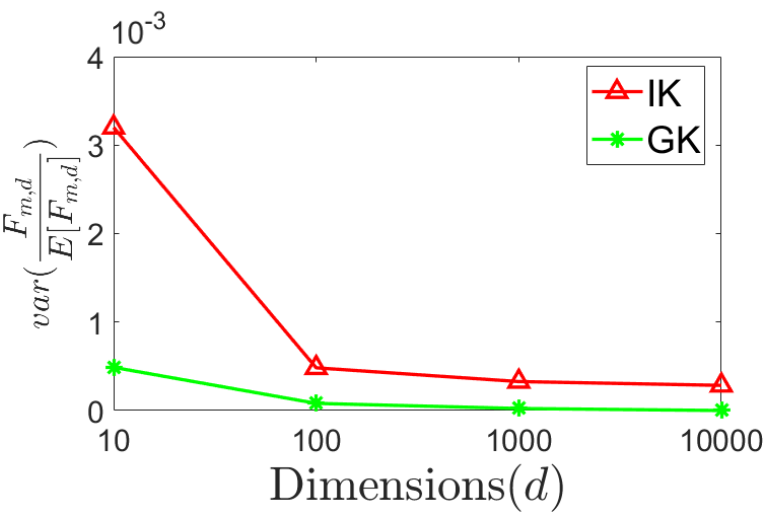}
			\caption{Concentration effect}
		\end{subfigure} 
		\begin{subfigure}[b]{0.245\textwidth}
			\centering
			\includegraphics[width=1.6in]{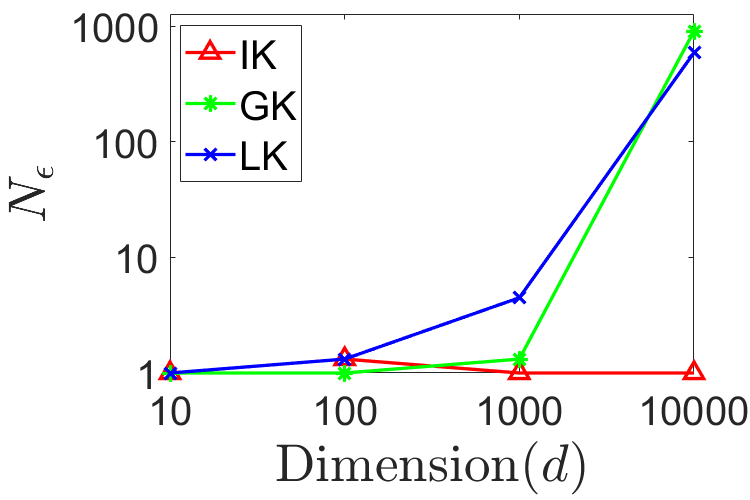}
			\caption{${\bf q}_i$: between 2 clusters}
		\end{subfigure}  
		\begin{subfigure}[b]{0.245\textwidth}
			\centering
			\includegraphics[width=1.6in]{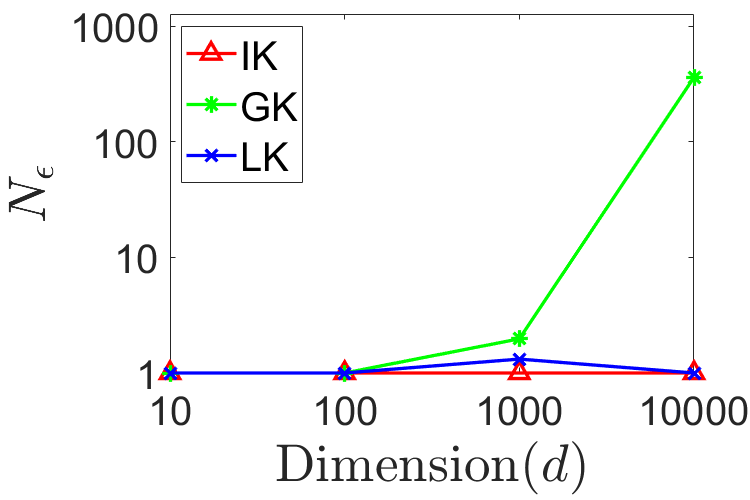}
			\caption{${\bf q}_c$: sparse cluster center}
		\end{subfigure} 
		
		\caption{The instability of a measure $\kappa$: Gaussian kernel (GK) and Linear kernel (LK) versus IK. The dataset used has 1,000 data points per cluster: the number of dimensions is increased from 10 to 10,000. 
			Query ${\bf q}_c$ is used in subfigure (b).
			$\epsilon=0.005$ is used in  subfigures (c) \& (d). 
		}
		\label{fig_std-mean} 
	\end{figure*}
	
	\begin{table*}[t]
		\caption{The instability of a measure.
			The dataset shown in Figure \ref{fig_std-mean}(a) is used here. $\epsilon=0.005$.} 
		\label{tbl_SNN_AG_IK} 
		\centering
		\begin{tabular}{cc|cc}
			\hline
			\multicolumn{2}{c|}{SNN, AG and IK} & \multicolumn{2}{c}{$\ell_p$-norm for $p=0.1,0.5$ \& $2$ } \\ \hline
			${\bf q}_i$: between 2 clusters & ${\bf q}_c$: sparse cluster center & ${\bf q}_i$: between 2 clusters & ${\bf q}_c$: sparse cluster center\\ \hline
			\includegraphics[width=1.3in]{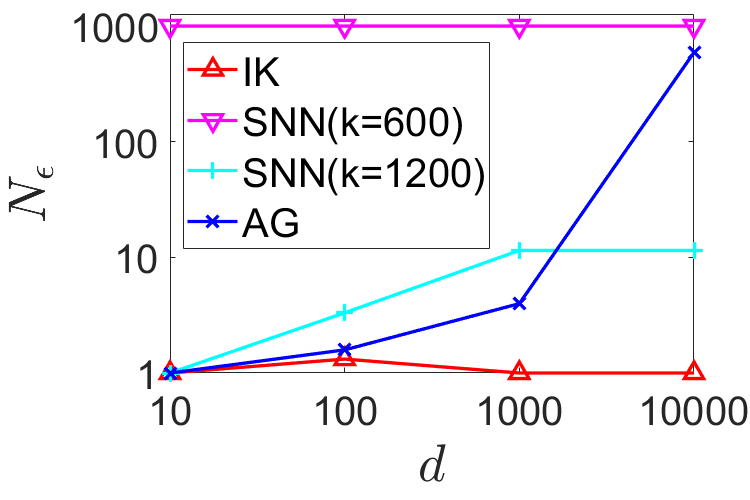} &
			\includegraphics[width=1.3in]{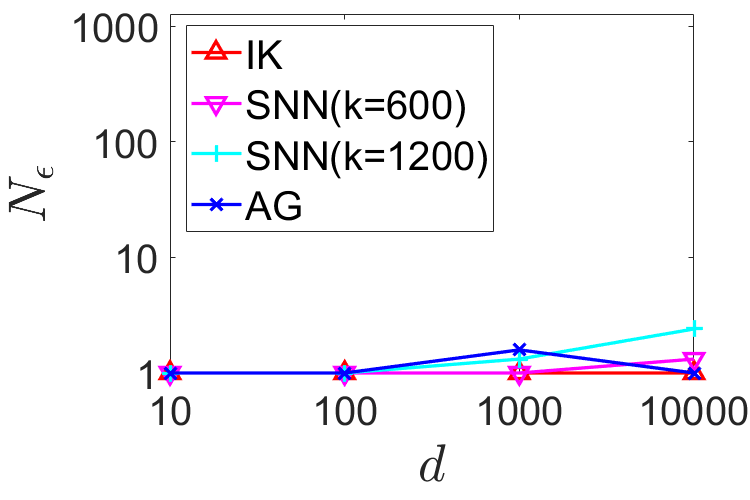} &
			\includegraphics[width=1.3in]{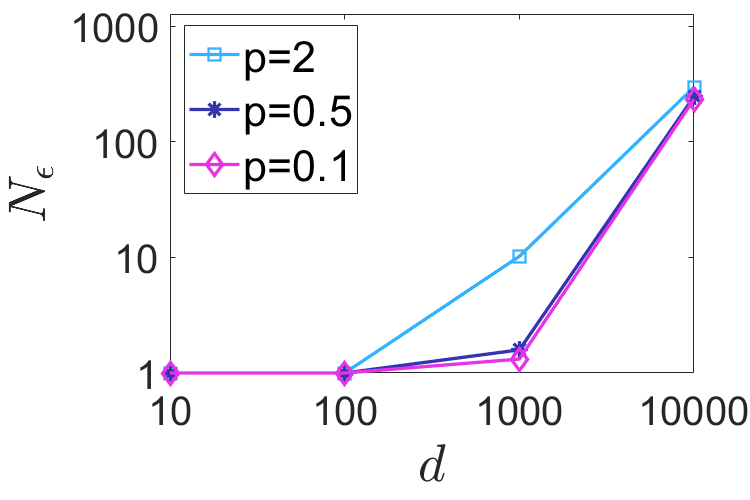} &    \includegraphics[width=1.3in]{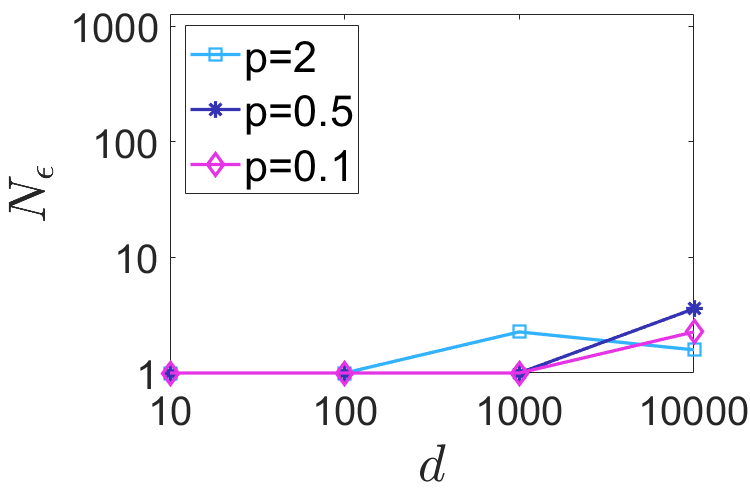} 
			\\\hline

		\end{tabular}
	\end{table*}

	\subsection{Instability of a measure}
	\label{sec_instability}
	The concentration effect is assessed in terms of the variance of measure distribution $F_{m,d}$  wrt the mean of the measure, i.e., $var \left( \frac{F_{m,d}}{\mathbb{E}[F_{m,d}]} \right)$, as the number of dimensions $d$ increases,
	based on a measure $m({\bf x},{\bf y}) = 1 - \kappa({\bf x},{\bf y})$.
	A measure is said to have the concentration effect if the variance of the measure distribution in dataset $D$ approaches zero as $d \rightarrow \infty$. 
	
	Following \cite{Beyer:1999}, we use the same notion of instability of a measure, as a result of a nearest neighbor query, to show the concentration effect. Let $m({\bf q}|D) = \min_{{\bf y} \in D} 1- \kappa({\bf q},{\bf y})$ be the distance as measured by $\kappa$ from query ${\bf q}$ to its nearest neighbor in dataset $D$; and $N_\epsilon$ be the number of points in $D$ having distances wrt ${\bf q}$ less than $(1+\epsilon)m({\bf q}|D)$.
	
	When there are many close nearest neighbors for any small $\epsilon>0$, i.e., $N_\epsilon$ is large, then the nearest neighbor query is said to be unstable. An unstable query is a direct consequence of the concentration effect. 
	Here we use $N_\epsilon$ as a proxy for indistinguishability: for small $\epsilon$, high $N_\epsilon$ signifies indistinguishability; and $N_\epsilon=1$ or close to 1 denotes distinguishability.

	Figure \ref{fig_std-mean} shows the verification outcomes using the dataset shown in Figure \ref{fig_std-mean}(a). 
	Figure \ref{fig_std-mean}(b) shows the concentration effect, that is, $var \left( \frac{F_{m,d}}{\mathbb{E}[F_{m,d}]} \right) \rightarrow 0$ as $d \rightarrow \infty$ for Gaussian kernel (GK). Yet, using Isolation Kernel (IK), the variance approaches a non-zero constant under the same condition. 
	
	Figures \ref{fig_std-mean}(c) \& \ref{fig_std-mean}(d) show that $N_\epsilon$ increases as $d$ increases for GK for two different queries; and LK is unstable in one query. Yet, IK has almost always maintained a constant $N_\epsilon$ ($=1$) as $d$ increases.
	
	
	
	
	\textbf{SNN, AG, Linear kernel and fractional distances}: 
	Here we examine the stability of Shared Nearest Neighbor (SNN) \citep{SNN,SNN-Defeat-Curse?}, Adaptive Gaussian kernel (AG) \citep{zelnik2005self} and fractional distances \cite{Aggarwal-2001}.

	Our experiment using SNN has shown that it has query stability if the query is within a cluster; but it has the worst query instability when the query is outside any clusters when $k$ is set less than the cluster size (1000 in this case), even in low dimensions!
	This result is shown in the first two subfigures in Table~\ref{tbl_SNN_AG_IK}. AG ($k=200$) has a similar behaviour as SNN that uses $k>1000$.

	The last two subfigures in Table \ref{tbl_SNN_AG_IK} show a comparison between Euclidean distance and fractional distances ($\ell_p, p=0.1,0.5$). It is true that the fractional distances are better than Euclidean distance in delaying the concentration effect up to $d=1000$; but they could not resist the concentration effect for higher dimensions.

	\textbf{Summary}. Many existing measures have indistinguishability in high dimensions that have prevented them from finding the  nearest neighbor if the query is outside of any clusters.
	\textbf{Only IK has distinguishability}. Note that IK uses distance to determine the (Voronoi) partition into which a point falls; but IK's similarity does not directly rely on distance. IK has a unique $t$-dimensional feature map that has increased distinguishability as $t$ increases, described in Section~\ref{sec_distinguishability}. No existing metric-based Lipschitz continuous kernels have a feature map with this property.

	\subsection{The impact of (in)distinguishability on four traditional tasks}
	\label{sec_traditional_tasks}
	
	
	We use the same ten datasets in the following four tasks. All have been used in previous studies (see Appendix \ref{ap1}), except two synthetic datasets: Gaussians and $w$-Gaussians. The former has two well-separated Gaussians, both on the same 10000 dimensions; and the latter has two $w$-dimensional Gaussians which overlap at the origin only, as shown in Figure \ref{fig_SC}. Notice below that algorithms which employ distance or Gaussian kernel have problems with the $w$-Gaussians dataset in all four tasks; but they have no problems with the Gaussians dataset.

	\begin{table*}[t]
		\centering
		\caption{Runtimes of exact 5-nearest-neighbors search.  Brute force vs Ball tree index. Boldface indicates faster runtime between brute force and index; or better precision.  The experimental details are in Appendix~\ref{ap1}. The Gaussians dataset is as used in Figure \ref{fig_std-mean}; $w$-Gaussians (where $w=5,000$) is as shown in Table \ref{fig:t-SNE}. The last two columns show the retrieval precision of 5 nearest neighbours. Every point in a dataset is used as a query; and the reported result is an average over all queries in the dataset.} 
		\begin{tabular}{|c|rrr|rr|rr|r|rr||rr|}
			\hline
			Dataset & $\#points$ & $\#dimen.$ & \#class & \multicolumn{2}{c|}{Distance} & \multicolumn{3}{c|}{IK} & \multicolumn{2}{c||}{LK} & \multicolumn{2}{c|}{Precision@5} \\\cline{5-13}
			& & & & Brute  & Balltr & Brute  & Balltr & Map &  Brute  & Balltr & Distance & IK\\
			\hline 
			Url & 3,000 & 3,231,961 & 2 & \textbf{38,702} &  41,508  & 112  & \textbf{109}  &  9 & \textbf{39,220} & 43,620 & .88  & \textbf{.95}\\ 
			News20 & 10,000 & 1,355,191 & 2
			& \textbf{101,243} & 113,445 & 3,475 & \textbf{2,257}  & 16 & \textbf{84,986} & 93,173 & .63  & \textbf{.88}\\
			Rcv1 & 10,000	& 47,236 & 2 & \textbf{3,615} & 4,037 & 739 & \textbf{578} & 11 & \textbf{5,699} & 6,272 & .90 & \textbf{.94}\\ 
			Real-sim &	10,000	&	20,958  & 2 &  \textbf{2,541}  & 2,863  & 4,824  &	\textbf{4,558}  & 21 & \textbf{2,533} & 2853 & .50 & \textbf{.63}\\ 
			
			Gaussians & 2,000 & 10,000 & 2 &  46  & \textbf{41} &  229 & \textbf{213} & 46 & \textbf{45} & 54 & \textbf{1.00} & \textbf{1.00} \\ 
			$w$-Gaussians & 2,000 & 10,000 & 2 &  \textbf{53} &  77 &    210  &  \textbf{205}  & 45  & \textbf{56}   &   73  & .90 & \textbf{1.00} \\     
			Cifar-10 &	10,000	&	3,072 & 10 &  \textbf{340} & 398 &  7,538 &	\textbf{7,169} & 67 & \textbf{367} & 439 & .25 & \textbf{.27}\\ 
			
			Mnist	&	10,000	&	780 & 10 & \textbf{58}   &  72   & 1,742 & \textbf{1,731} & 2 & \textbf{87} & 106 & \textbf{.93} & \textbf{.93} \\
			A9a	&	10,000	&	122 & 2 & \textbf{10}    &	13 & 5,707 & \textbf{5,549} & 1 & \textbf{12} & 16 & \textbf{.79} & \textbf{.79} \\ 
			Ijcnn1 & 10,000 & 22 & 2 & 2.3 & \textbf{1.3} & 706  &	\textbf{654} & 1 & 2.2 & \textbf{1.8} & \textbf{.97} & .96\\
			\hline
			\multicolumn{11}{|r}{\textit{Average}}     & .77  & \textbf{.84}  \\
			\hline
		\end{tabular}%
		\label{tbl_indexing}%
	\end{table*}%
	

	\subsubsection{Exact nearest neighbor search using ball tree indexing}
	\label{sec_ballTree}
	

	Existing indexing techniques are sensitive to dimensionality, data size and structure of the dataset \citep{NN-indexibility-TODS2006,ExactNNSearch-PKDD2007}. They only work in low dimensional datasets of a moderate size where a dataset has clusters. 


	Table \ref{tbl_indexing} shows the comparison between the brute force search and an indexed search based on ball tree \citep{omohundro1989five}, in terms of runtime, using Euclidean distance, IK and linear kernel (LK). 
	Using LK, the ball tree index is always worse than the brute force search in all datasets, except the lowest dimensional dataset. Using Euclidean distance, we have a surprising result that the ball tree index ran faster than the brute force in one high dimensional dataset. However, this result cannot be `explained away' by using the argument that they may have low intrinsic dimensions (IDs) because the Gaussians dataset has 10,000 (theoretically true) IDs; while the indexed search ran slower in  all other real-world high dimensional datasets where each has the median ID less than 25, as estimated using a recent local ID estimator \citep{LID-SIAM19} (see the full result in Appendix \ref{sec_ID}.)

	
	Yet, with an appropriate value of $\psi$, IK yielded a faster search using the ball tree index than the brute force search in all datasets, without exception. Note that the comparison using IK is independent of the feature mapping time of IK because both the brute force and the ball tree indexing require the same feature mapping time (shown in the `Map' column in  Table \ref{tbl_indexing}.) 

	It is interesting to note that the feature map of IK uses an effective number of dimensions of $t=200$ in all experiments. While one may tend to view this as a dimension reduction method, it is a side effect of the kernel, not by design. The key to IK's success in enabling  efficient indexing is the kernel characteristic in high dimensions (in Hilbert space.) IK is not designed for dimension reduction. 
	
	Note that IK works not because it has used a low-dimensional feature map. Our result in Table \ref{tbl_indexing} is consistent with the previous evaluation (up to 80 dimensions only \citep{ExactNNSearch-PKDD2007}), i.e.,  indexing methods ran slower than brute-force when the number of dimensions is more than 16. If the IK's result is due to low dimensional feature map, it must be about 20 dimensions or less for IK to work. But all IK results in Table \ref{tbl_indexing} have used $t=200$. 
	
	Nevertheless, the use of $t=200$ effective dimensions has an impact on the absolute runtime. Notice that, when using the brute force method, the actual runtime of IK  was one to two orders of magnitude shorter than that of distance in $d>40,000$; and the reverse is true in $d<40,000$.
	We thus recommend using IK in high dimensional ($d>40,000$) datasets, not only because its indexing runs faster than the brute force, but IK runs significantly faster than distance too. In $d\le 40,000$ dimensional datasets, distance is preferred over IK because the former runs faster. 
	
	
	The last two columns in Table \ref{tbl_indexing} show the retrieval result in terms of precision of 5 nearest neighbors. It is interesting to note that IK has better retrieval outcomes than distance in all cases, where large differences can be found in News20, Realsim and $w$-Gaussians. The only exception is Ijcnn1 which has the lowest number of dimensions, and the difference in precision is small.
	
	This shows that the concentration effect influences not only the indexing runtime but also the precision of retrieval outcomes.
	
	\subsubsection{Clustering}
	\label{sec_SC}
	
	In this section, we examine the effect of IK versus GK/distance using a scalable spectral clustering algorithm (SC) \citep{chen2010parallel}  and Density-Peak clustering (DP) \citep{DP2014clustering}.
	We report the average clustering performance in terms of AMI (Adjusted Mutual Information) \citep{vinh2010information} of SC over 10 trials on each dataset because SC is a randomized algorithm; but only a single trial is required for DP, which is a deterministic algorithm.

	\begin{table}[h]
		\centering
		\caption{Best AMI scores of Desity-Peak clustering (DP) and Spectral Clustering (SC) using Distance/Gaussian kernel (GK), Adaptive Gaussian kernel (AG) and IK. Note that every cluster in Cifar10 and Ijcnn1 has mixed classes. That is why all clustering algorithms have very low AMI.}
		\begin{tabular}{|c|ccc|ccc|}
			\hline
			Dataset &   \multicolumn{3}{c|}{DP} & \multicolumn{3}{c|}{SC} \\
			\cline{2-7}
			& Distance & AG    & IK    & GK & AG    & IK \\
			\hline
			Url   & .07  & \textbf{.19}  & .16 & .04  & .04  & \textbf{.07}\\
			News20  & .02  & .05  & \textbf{.27} & .16  & .15  & \textbf{.21} \\
			Rcv1  & .18  & .19  & \textbf{.32} & \textbf{.19} & \textbf{.19} & \textbf{.19} \\
			Real-sim & .02  & .03  & \textbf{.04} & .04  & .02  & \textbf{.05} \\
			Gaussians & \textbf{1.00} & \textbf{1.00} & \textbf{1.00} & \textbf{1.00} & \textbf{1.00} & \textbf{1.00} \\
			$w$-Gaussians & .01  & .01  & \textbf{.68} & .24  & .25  & \textbf{1.00} \\
			Cifar-10 & .08  & .09  & \textbf{.10} & \textbf{.11} & \textbf{.11} & .10 \\
			Mnist & .42  & \textbf{.73} & .69  & .49  & .50  & \textbf{.69} \\
			A9a   & .08  & .17  & \textbf{.18} & .15  & .14  & \textbf{.17} \\
			Ijcnn1 & .00  & .00  & \textbf{.07} & \textbf{.00} & \textbf{.00} & \textbf{.00} \\
			\hline
			\textit{Average} & .19  & .25  & \textbf{.35}  & .24  & .24  & \textbf{.35}\\
			\hline
		\end{tabular}%
		\label{tbl_SC2}%
	\end{table}%

	\begin{table*}[t]
		\centering
		\caption{SVM classification accuracy \& runtime (CPU seconds). Gaussian kernel (GK), IK and Linear kernel (LK). $nnz\% = \#nonzero\_values / ((\#train+\#test) \times \#dimensions) \times 100$. 
			IK ran five trials to produce [mean]{\tiny $\pm$}[standard error]. The runtime of SVM with IK includes the IK mapping time. The separate runtimes of IK mapping and SVM are shown in Appendix \ref{sec_IK_mapping_time}. 
			\label{tbl:SVM-accuracy} }

		\begin{tabular}{|c|rrrr|rrr||rrr|}\hline
			Dataset		& $\#train$     & $\#test$ & $\#dimen.$ & $nnz\%$ & \multicolumn{3}{c||}{Accuracy} & \multicolumn{3}{c|}{Runtime}\\
			& & & & 	& LK & \multicolumn{1}{c}{IK} & GK & LK & IK & GK\\
			\hline
			Url &     2,000 & 1,000 & 3,231,961 & .0036 & \textbf{.98} & \textbf{.98}{\tiny $\pm$.001} & \textbf{.98} & 2 & 3 & 14 \\
			News20 & 15,997 & 3,999 & 1,355,191 & .03 
			& .85 & \textbf{.92}{\tiny $\pm$.007} & .84  & 38 & 60 & 528 \\
			Rcv1 & 20,242 & 677,399 & 47,236 & .16 
			& .96  & .96{\tiny $\pm$.013} & \textbf{.97} & 111 & 420 & 673\\ 
			Real-sim & 57,848 & 14,461 & 20,958 & .24 
			& \textbf{.98} & \textbf{.98}{\tiny $\pm$.010} & \textbf{.98} & 49 & 57 & 2114\\ 
			Gaussians & 1,000 & 1,000 & 10,000 & 100.0 &  \textbf{1.00} & \textbf{1.00}{\tiny $\pm$.000} & \textbf{1.00} & 14 & 9 & 78\\ 
			$w$-Gaussians & 1,000 & 1,000 & 10,000 & 100.0  & .49 & \textbf{1.00}{\tiny $\pm$.000} & .62 & 20 & 8 & 79  \\
			Cifar-10 & 50,000 & 10,000 & 3,072 & 99.8 
			& .37 & \textbf{.56}{\tiny $\pm$.022} & .54 & 3,808 & 1087 & 29,322\\ 
			Mnist & 60,000 & 10,000 & 780 & 19.3 
			& .92 & .96{\tiny $\pm$.006} & \textbf{.98} & 122 &  48  & 598\\
			A9a & 32,561 & 16,281 & 123 & 11.3 
			& \textbf{.85} & \textbf{.85}{\tiny $\pm$.012} & \textbf{.85} & 1 & 31 & 100\\
			Ijcnn1 & 49,990 & 91,701 & 22 & 59.1 
			& .92 & .96{\tiny $\pm$.006} & \textbf{.98} & 5 & 66 & 95\\
			\hline
		\end{tabular}
	\end{table*}
	

	Table \ref{tbl_SC2} shows the result. 
	With SC, IK improves over GK or AG \citep{zelnik2005self} in almost all datasets. Large improvements are found on $w$-Gaussians and Mnist. The only dataset in which distance and GK did well is the Gaussians dataset that has two well-separate clusters.
	
	AG was previously proposed to `correct' the bias of GK \citep{zelnik2005self}. However, our result shows that AG is not better than GK in almost all datasets we have used. There are two reasons. First, we found that previous evaluations have searched for a small range of $\sigma$ of GK (e.g., \citep{chen2010parallel}.) Especially in high dimensional datasets, $\sigma$ shall be searched in a wide range (see Table \ref{tbl_SC_parameters} in Appendix \ref{ap1}.) Second, spectral clustering is weak in some data distribution even in low dimensional datasets. In this case, neither GK nor AG makes a difference. This is shown in the following example using $w$-Gaussians, where $w=1$, which contains two 1-dimensional Gaussian clusters which only overlap at the origin in the 2-dimensional space.

\begin{figure}[h]
		\centering 
		\begin{subfigure}[b]{0.22\textwidth}
			\centering
			\includegraphics[width=1.4in]{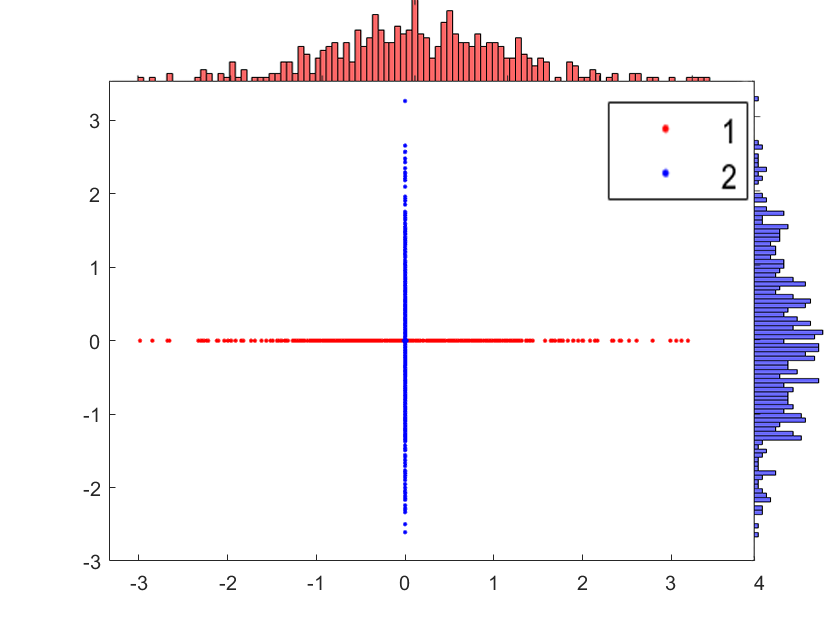}
			\caption{SC with IK (AMI=0.91)}
		\end{subfigure}
		\begin{subfigure}[b]{0.22\textwidth}
			\centering
			\includegraphics[width=1.4in]{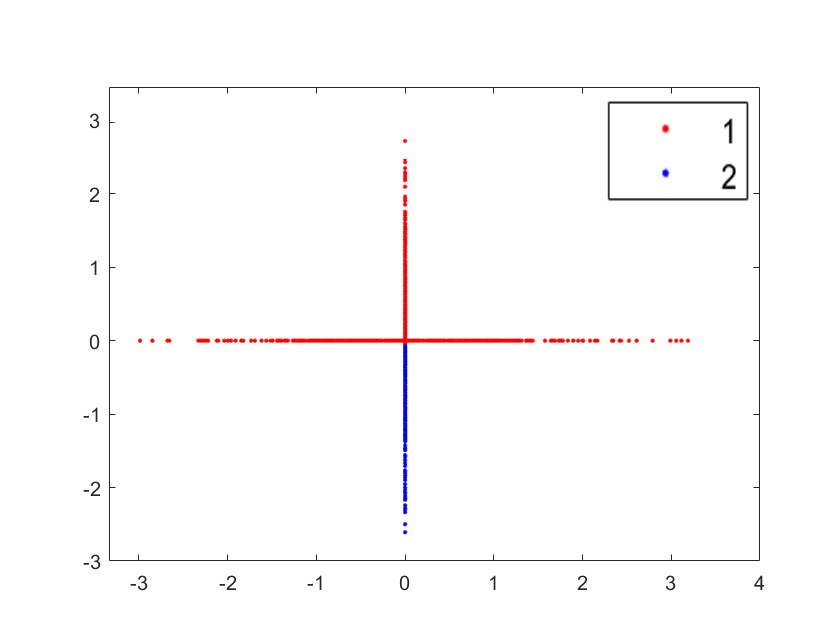}
			\caption{SC with AG (AMI=0.29)}
		\end{subfigure}  
		\caption{The spectral clustering results with IK and AG on the $w$-Gaussians dataset (where $w=1$) that contains two 1-dimensional subspace clusters. Each cluster has 500 points, sampled from a $1$-dimensional Gaussian distribution $\mathcal{N}(0, 1)$. The density distribution of each cluster is shown in (a) wrt the corresponding 1-dimensional subspace. }
		\label{fig_SC}
	\end{figure}

	Figure \ref{fig_SC} shows the best spectral clustering results with IK and AG.
	When the two Gaussians in the $w$-Gaussians dataset have the same variance, spectral clustering with AG (or GK) fails to separate the two clusters; but spectral clustering with IK succeeds. 
	
	When the variances of the two subspace clusters are substantially different,  spectral clustering with AG yields a comparable good result to that with IK. Note that spectral clustering with GK cannot separate the two subspace clusters regardless. This is the specific condition, as used previously \citep{zelnik2005self} to show that AG is better than GK in spectral clustering.

	In a nutshell, in general, IK performs better than GK and AG on spectral clustering in both low and high dimensional datasets.

	This clustering result is consistent with the previous result \citep{IsolationKernel-AAAI2019} conducted using DBSCAN \citep{ester1996density} in both low and high dimensional datasets comparing between distance and IK.
	
	Similar relative results with DP \cite{DP2014clustering} are shown in Table \ref{fig_SC}, where large differences in AMI in favor of IK over distance/AG can be found in News20 and $w$-Gaussians.

	\begin{table*}[t]
		\caption{t-SNE: Gaussian Kernel versus Isolation Kernel on the $w$-Gaussians datasets having two $w$-dimensional subspace clusters (shown in the first column.)  Both clusters are generated using $\mathcal{N}(0, 1)$.  ``$\times$'' is the origin---the only place in which the two clusters overlap in the data space. }
		\label{fig:t-SNE}
		\centering
		\begin{tabular}{c|ccccc}
			\hline
			$w=1$ dataset  &  &  t-SNE(GK)   & t-SNE(IK)  &  FIt-SNE  & q-SNE\\
			\hline 
			\includegraphics[width=1.2in]{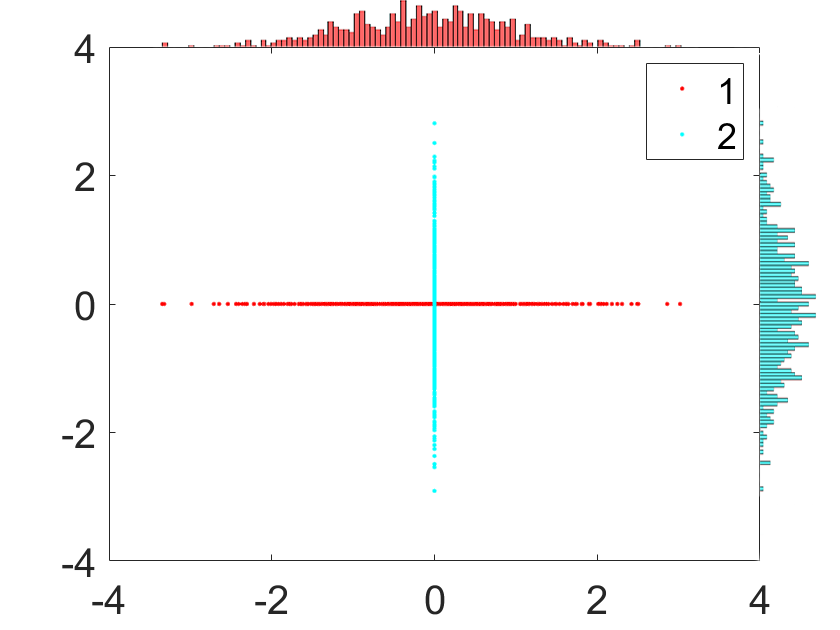} &       \begin{turn}{90} \quad   $w=2$   \end{turn}&       
			\includegraphics[width=1.2in]{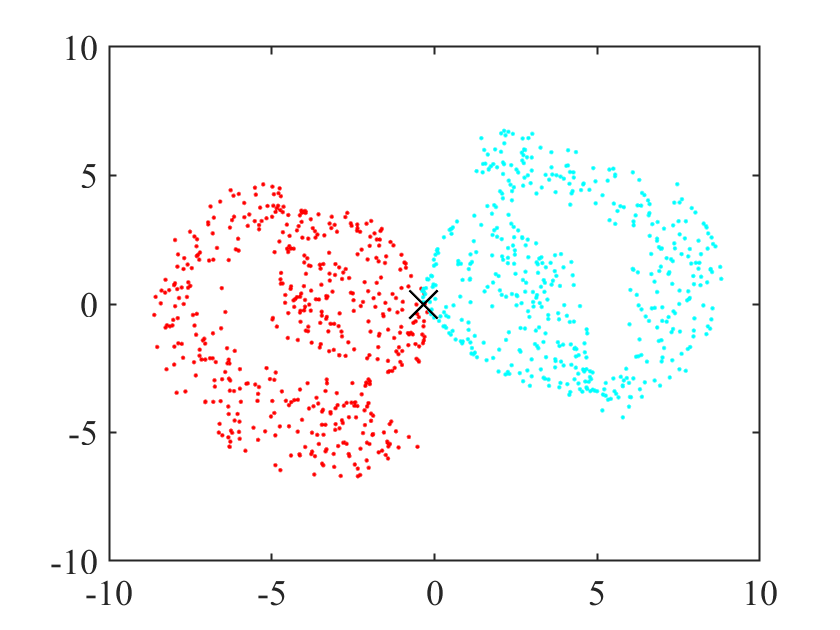}                     &
			\includegraphics[width=1.2in]{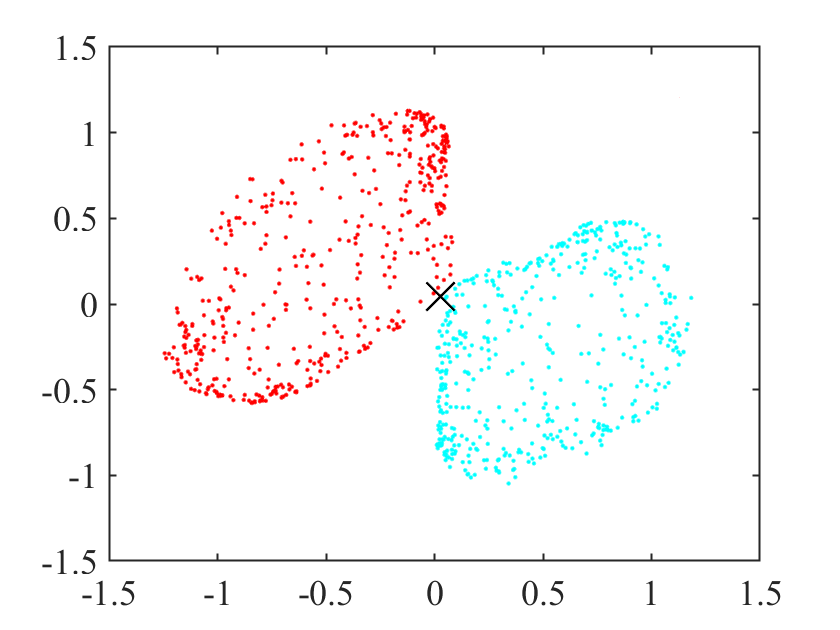}  &
			\includegraphics[width=1.2in]{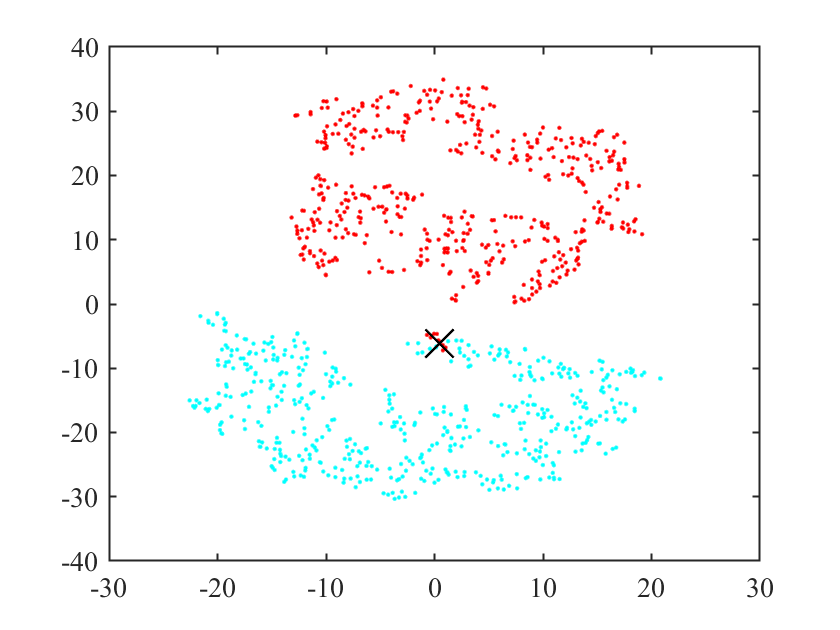}  &
			\includegraphics[width=1.2in]{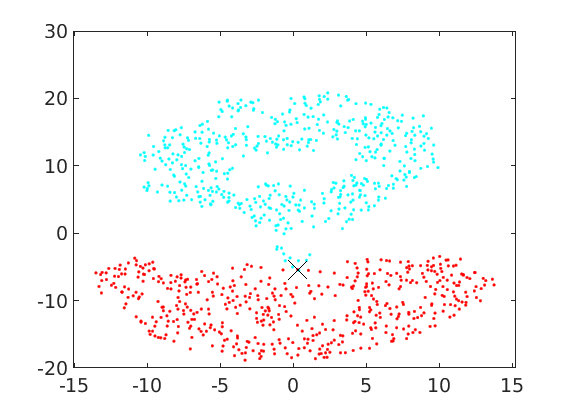}\\  
			&   \begin{turn}{90} \quad  $w=5000$ \end{turn}&       
			\includegraphics[width=1.2in]{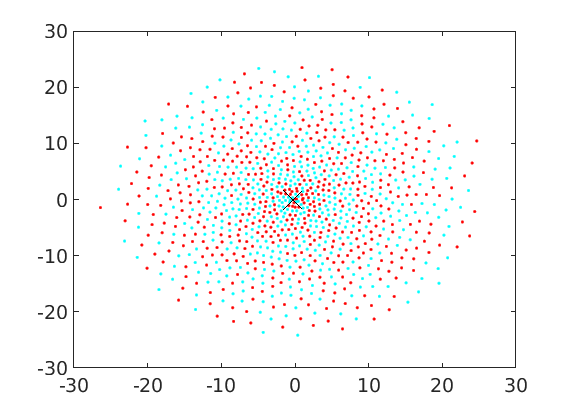}                           &
			\includegraphics[width=1.2in]{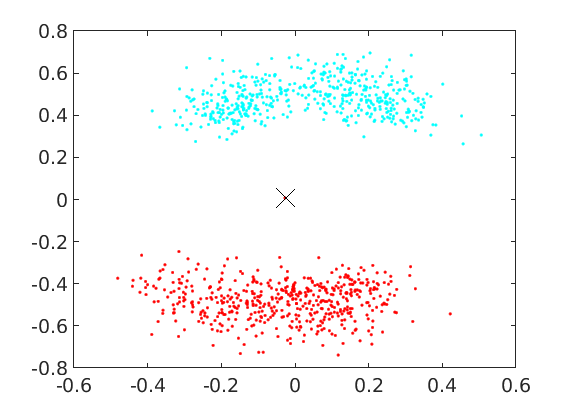}  &
			\includegraphics[width=1.2in]{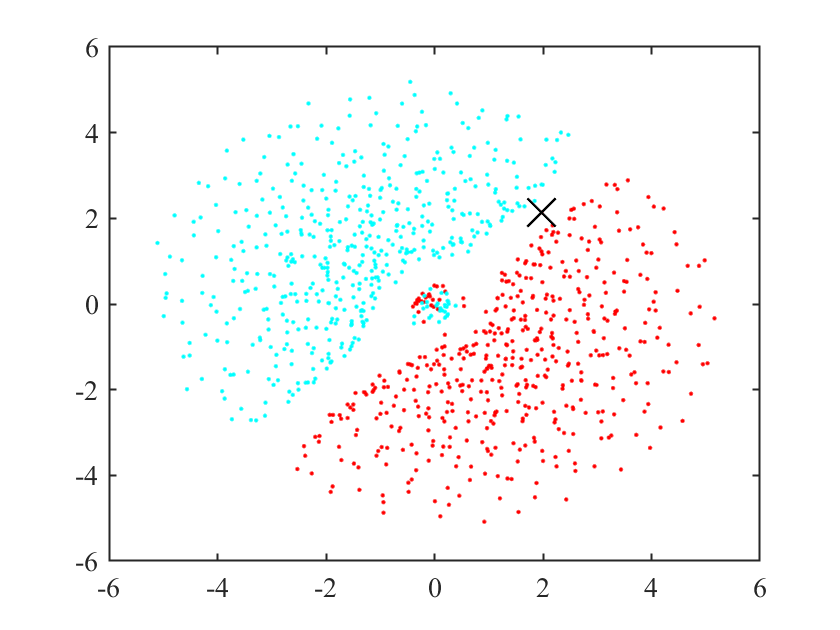} &
			\includegraphics[width=1.2in]{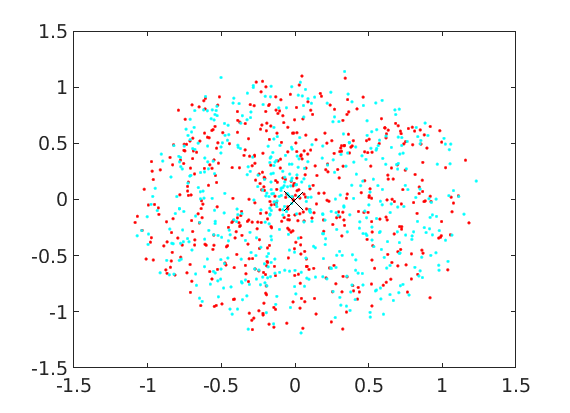} \\     
			\hline
		\end{tabular} 
	\end{table*}
	
	\begin{table*} [t]
		\caption{t-SNE: GK versus IK on $w$-Gaussians datasets having two $w$-dimensional subspace clusters (different variances.)  The red and blue clusters are generated using $\mathcal{N}(0, 1)$ and $\mathcal{N}(0, 32)$, respectively. 
			``$\times$'' is the origin---the only place in which the two clusters overlap in the data space.  }
		\label{fig:t-SNE-different-variance}
		\centering
		\begin{tabular}{c|ccccc}
			\hline
			$w=1$ dataset  &  &  t-SNE(GK)   & t-SNE(IK)  &  FIt-SNE  & q-SNE\\
			\hline 
			\includegraphics[width=1.2in]{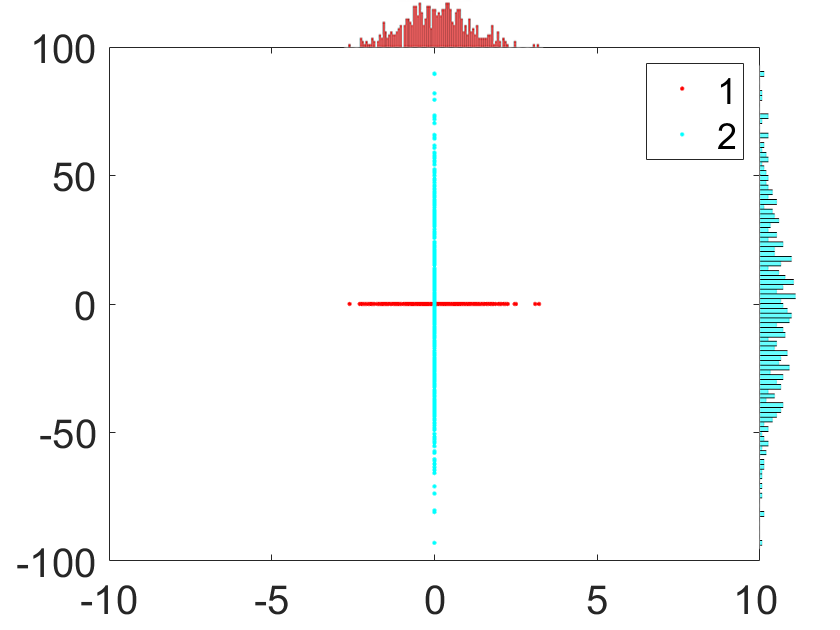} &       \begin{turn}{90} \quad   $w=2$   \end{turn}&      
			\includegraphics[width=1.2in]{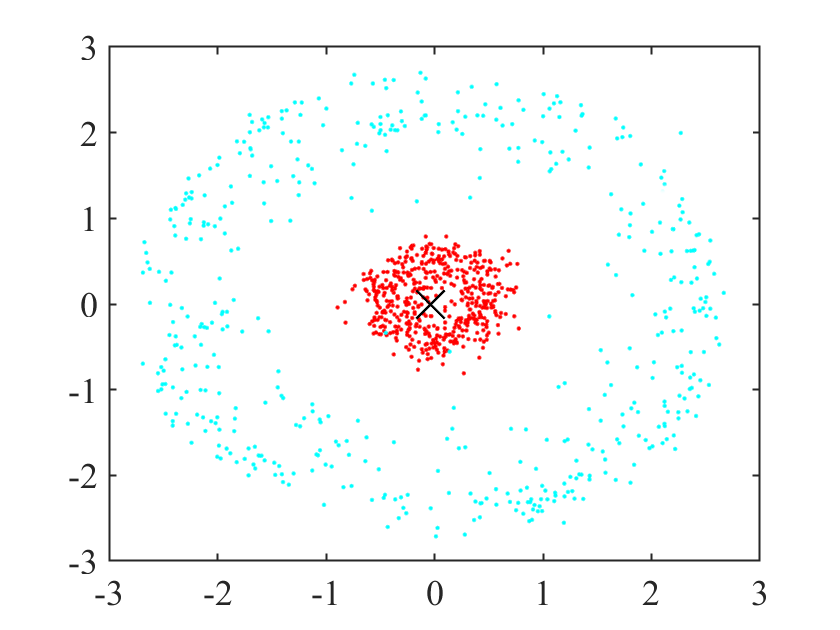}                     &
			\includegraphics[width=1.2in]{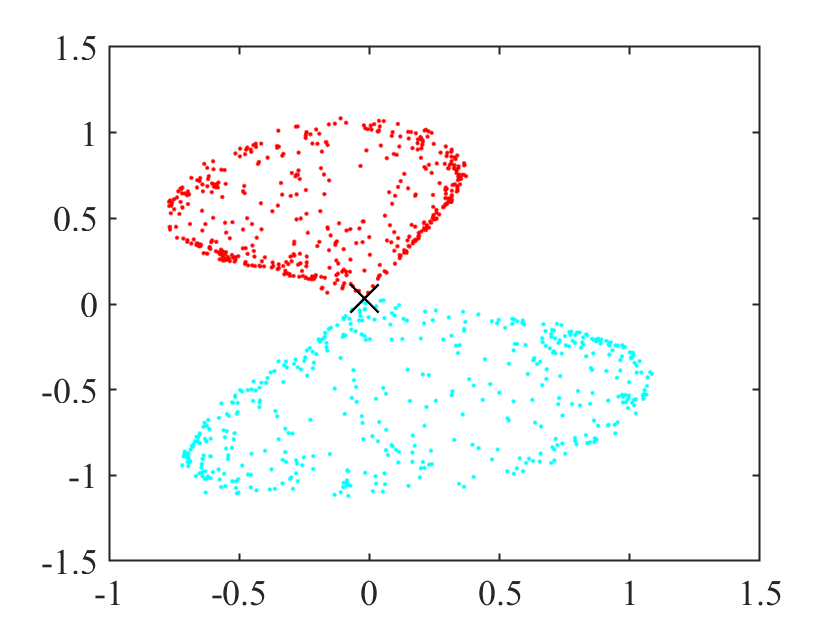}      &
			\includegraphics[width=1.2in]{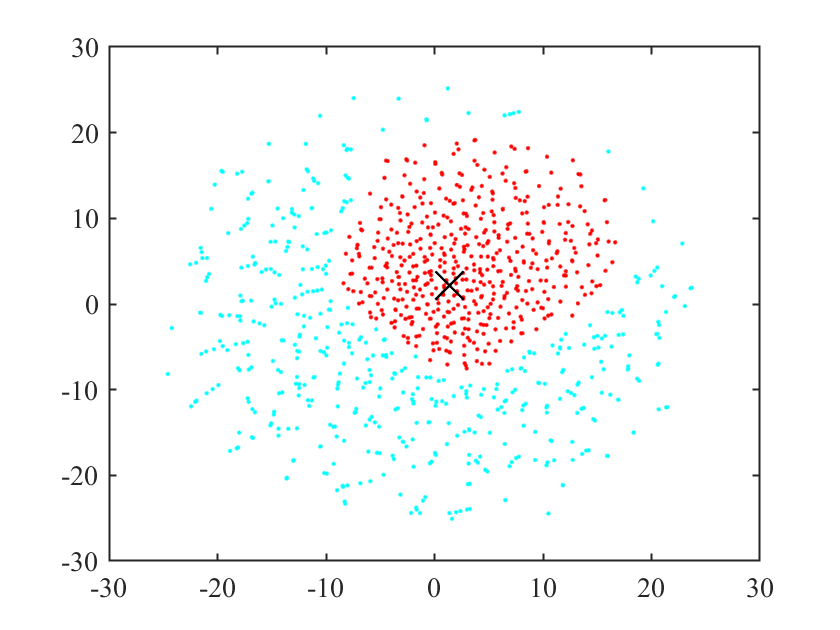} &  
			\includegraphics[width=1.2in]{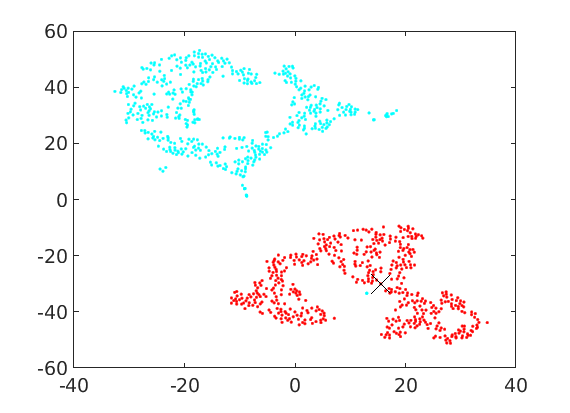} \\ 
			&     \begin{turn}{90}    \quad  $w=100$ \end{turn}&       
			\includegraphics[width=1.2in]{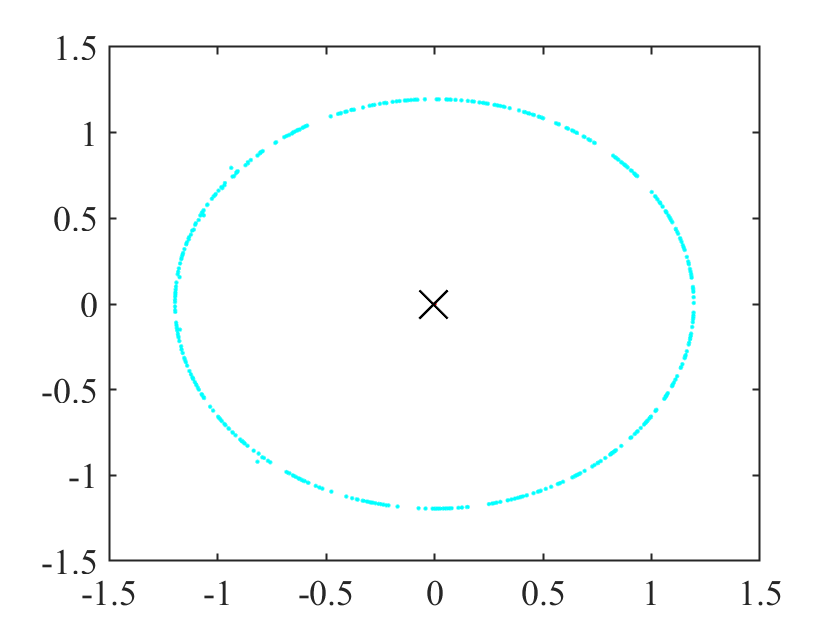}                           &
			\includegraphics[width=1.2in]{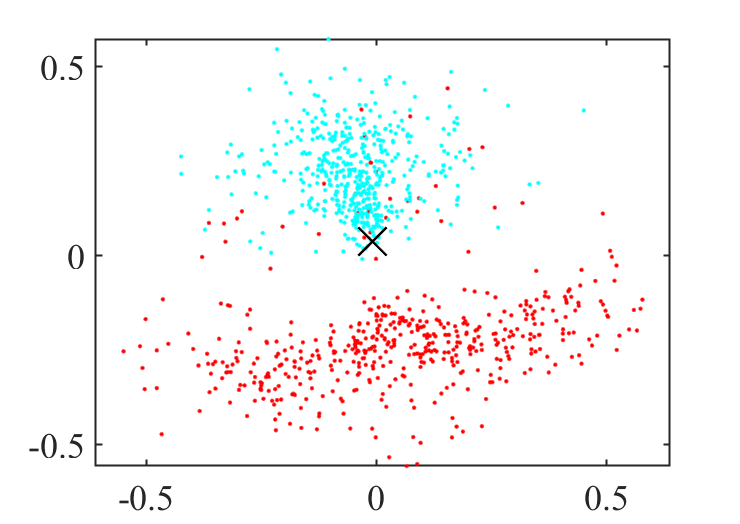}     &
			\includegraphics[width=1.2in]{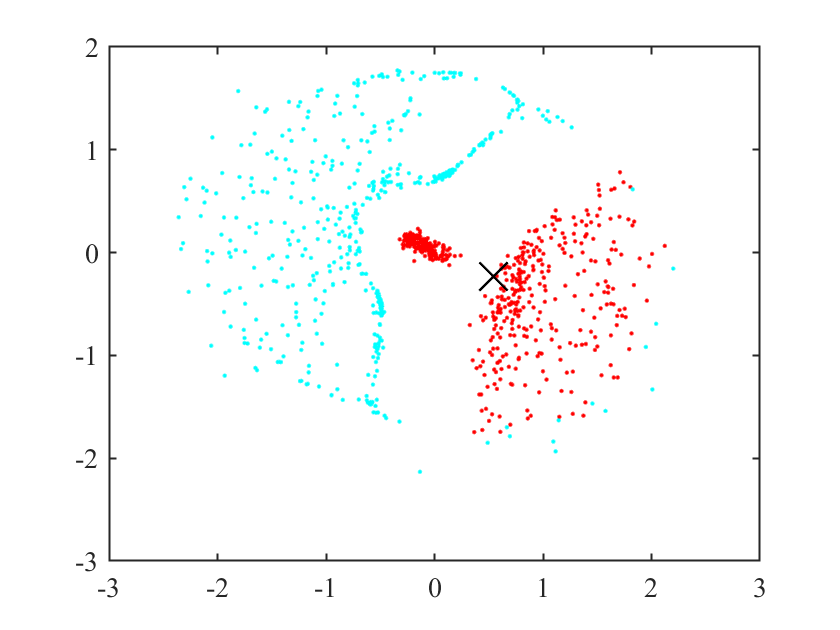}  &    
			\includegraphics[width=1.2in]{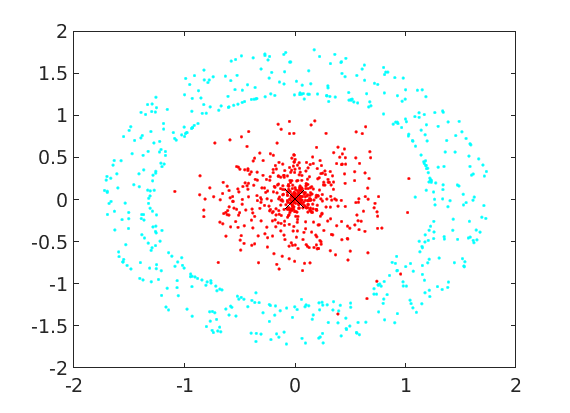} \\     
			\hline
		\end{tabular}
	\end{table*}

	\subsubsection{SVM classification}
	\label{sec_SVM-classification}
	Table \ref{tbl:SVM-accuracy} shows the comparison between linear kernel, IK and Gaussian kernel using an SVM classifier in terms of accuracy and runtime. The experimental details are in Appendix~\ref{ap1}.

	It is interesting to note that IK produced consistently high (or the highest) accuracy in all datasets. In contrast, both linear and Gaussian kernels have a mixed bag of low and high accuracies. For example, they both perform substantially poorer than IK on News20 and $w$-Gaussians (where $w=5,000$.) Each $w$-Gaussians dataset has two $w$-dimensional subspace clusters. The $w=1$ version of the dataset is shown in the first column in Table \ref{fig:t-SNE}.
	
	The runtime result shows that,  SVM using IK runs either comparably or faster than SVM using Linear Kernel, i.e., they are in the same order, especially in high dimensions. Because of employing LIBLINEAR \citep{fan2008liblinear}, both are up to two orders of magnitude faster  in large datasets than SVM with GK which must employ the slower nonlinear LIBSVM \citep{CC01a}.
	
	
	\subsubsection{Visualization using t-SNE}
	\label{sec_visualization_experiment}
	
	
	Studies in visualization \citep{MDS-wickelmaier2003introduction,Hinton2008Visualizing} often ignore the curse of dimensionality issue which raises doubt about the assumption made by visualization methods. For example, t-SNE \citep{Hinton2008Visualizing} employs Gaussian kernel as a means to measure similarity in the high dimensional data space. No studies have examined the effect of its use in the presence of the curse of dimensionality, as far as we know.
	Here we show one example misrepresentation from t-SNE, due to the use of GK to measure similarity in the high dimensional data space on the $w$-Gaussians datasets.

	

	The t-SNE visualization results comparing Gaussian kernel and Isolation Kernel are shown in the middle two columns in Table \ref{fig:t-SNE}. While t-SNE using GK has preserved the structure in low dimensional  ($w=2$) data space, it fails completely to separate the two clusters  in high dimensional ($w=5,000$) data space. 
	In contrast, t-SNE with IK correctly separates the two clusters in both low and high dimensional data spaces.
	
	Recent improvements on t-SNE, e.g., FIt-SNE \citep{linderman2019fast} and q-SNE \citep{hakkinen2020qsne}, explore  better local structure resolution or efficient approximation. As they employ Gaussian kernel, they also produce misrepresented structures, as shown in Table \ref{fig:t-SNE}. 
	
	Table \ref{fig:t-SNE-different-variance} shows the results on a variant $w$-Gaussians dataset where the two clusters have different variances. Here t-SNE using GK produces misrepresented structure in both low and high dimensions. It has misrepresented the origin in the data space as belonging to the red cluster only, i.e., no connection between the two clusters. Note that all points in the red cluster are concentrated at the origin in $w=100$ for GK. There is no such issue with IK.

	
	
	
	In low dimensional data space, this misrepresentation is due solely to the use of a data independent kernel. To make GK adaptive to local density in t-SNE, the bandwidth is learned locally for each point. As a result, the only overlap between the two subspace clusters, i.e., the origin,  is assigned to the dense cluster only. 
	The advantage of IK in low dimensional data space is due to the data dependency stated in Lemma \ref{lem_characteristic}. This advantage in various tasks has been studied previously \citep{ting2018IsolationKernel,IsolationKernel-AAAI2019,IsolationSetKernel,IDK-KDD2020}.
	
	In high dimensional data space, the effect due to the curse of dimensionality obscures the similarity measurements made. Both this effect and the data independent effect collude in high dimensional data space when distance-based measures are used, leading to poor result in the transformed space. 
	
	\vspace{-1mm}
	\subsubsection{Further investigation using the low dimensional $w$-Gaussians ($w=2$)}
	It is interesting to note that (i) in SVM classification, both GK and IK could yield 100\% accuracy while LK still performed poorly at 51\% accuracy; and (ii) all three measures enable faster indexed search than brute force search and have equally high precision on this low dimensional dataset. But only IK can do well in the high dimensional $w$-Gaussians ($w=5,000$) in all four tasks, as shown in Tables \ref{tbl_indexing} to \ref{fig:t-SNE-different-variance}.   
	
	The $w$-Gaussians datasets provide an example condition in which GK/distance performed well in low dimensions but poorly in high dimensions in indexed search, SVM classification and t-SNE visualization.
	This is a manifestation that the GK/distance used in these algorithms has indistinguishability in high dimensions. 
	

	\noindent
	\textbf{Summary}: 
	\begin{itemize}[itemsep=0ex, leftmargin=5mm]
		\item IK is the only measure that consistently (a) enabled faster indexed search than brute force search, (b) provided good clustering outcomes with SC and DP,  (c) produced  high accuracy in SVM classification, and (d) yielded matching structures in t-SNE visualization in both low and high dimensional datasets.
		\item Euclidean distance or GK performed poorly in most high dimensional datasets in all tasks except SVM classification. We show one condition, using the artificial $w$-Gaussians datasets, in which they did well in low dimensions but performed poorly in high dimensions in all four tasks. The Gaussians dataset is one condition in which Euclidean distance or GK can do well in high dimensions. 
		\item Though low intrinsic dimensions may explain why GK/LK can do well in some high dimensional datasets in SVM classification, the explanation failed completely for all the slower indexed searches in high dimensions, and for the only faster indexed search on the Gaussians dataset as it has high intrinsic dimensions.  
	\end{itemize}

	

	\section{Discussion}
	\label{sec_discussion}
	We have limited our discussion on the curse of dimensionality based on nearest neighbors, i.e., their (in)distinguishability. But the effect of the curse is broader than this. For example, it has effects on the consistency of estimators, an aspect not discussed in this article.
	
	In addition to the concentration effect \citep{Talagrand,Beyer:1999}, the curse of dimensionality has also been studied under different phenomena, e.g., the correlation between attributes \citep{WhenNNisMeaningful-2009} and hubness \citep{HubsInSpace-JMLR-2010}.
	It is interesting to analyze whether IK can deal with these issues, other than the concentration effect, effectively in high dimensions.
	

	Our results in Tables \ref{tbl_indexing} to \ref{tbl:SVM-accuracy} show that some existing measures may perform well in high dimensions under special conditions. However, we are not aware of analyses that examine whether any existing measures can break the curse of dimensionality.
	
	
	Previous work on intrinsic dimensionality (e.g., \cite{LID-NIPS2011}) has shown that nonparametric distance-based regressors could escape the curse of dimensionality if a high dimensional dataset has low intrinsic dimensions. This is consistent with the finding that the concentration effect depends more on the intrinsic dimensions $I_d$ than on the input dimensions $d > I_d$ \citep{Concentration-Fractionaldistances}. Our result in Table \ref{tbl_indexing} on the Gaussians dataset adds another element that needs an explanation for existing measures, i.e., why does distance perform well on datasets with high intrinsic dimensions? 
	
	Because our analysis on IK's distinguishability is independent of dimensionality and data distribution, IK does not rely on low intrinsic dimensions to perform well, as shown in the artificial datasets: both Gaussians and $w$-Gaussians have high ($\ge 5,000$ true) intrinsic dimensions in Tables \ref{tbl_indexing} to \ref{fig:t-SNE-different-variance}.
	

	Our applications in ball tree indexing, two clustering algorithms, SVM classification and t-SNE here, and previously in DBSCAN clustering \citep{IsolationKernel-AAAI2019}  and multi-instance learning \cite{IsolationSetKernel}, suggest that many existing algorithms can get performance improvement by simply replacing distance/kernel with IK. However, a caveat is in order: not all existing algorithms can achieve that. For example, OCSVM \citep{OCSVM2001} and OCSMM \citep{OCSMM2013} have been shown to work poorly with IK \citep{ting2020-IDK-GroupAnomalyDetection}. This is because the learning in these two algorithms is designed to have a certain geometry in Hilbert space in mind; and IK does not have such a geometry (see details in \cite{ting2020-IDK-GroupAnomalyDetection}.) 
	
	
	
	The isolating mechanism used has a direct influence on IK's distinguishability. 
	Note that the 
	proofs of Lemma \ref{lem2} and Theorem \ref{thm2} rely on the isolating partitions being the Voronoi diagram.  It remains an open question whether IK with a different implementation could be proven to have distinguishability.
	
	Kernel functional approximation is an influential approach to obtain an approximate finite-dimensional feature map from an infinite-dimensional feature map of a kernel such as Gaussian kernel. Its representative methods are Nystr\"{o}m \citep{Nystrom_NIPS2000} and random features \citep{RandomFeatures2007,OrthonogalRandomFeatures2016}. The former is often dubbed data dependent and the latter data independent. These terms refer to the use (or no use) of data samples to derive the approximate feature map. In either case, the kernel employed is data independent and translation invariant, unlike the data dependent IK in which the similarity depends local data distribution and it is not translation invariant.
	
	VP-SVM \citep{VP-SVM-2016} employs Voronoi diagram to split the input space and then learns a local SVM with a local bandwidth of Gaussian kernel in each Voronoi cell. It aims to reduce the training time of SVM only, and do not address the curse of dimensionality at all.
	
	A recent work \citep{IK-SNE-2021} has already shown that IK improves the effectiveness and efficiency of t-SNE, by simply replacing Gaussian kernel. However, it did not address the issue of curse of dimensionality, which is the focus of this paper.

	The result of an additional investigation on the hubness effect \cite{HubsInSpace-JMLR-2010}  can be found in Appendix  \ref{App_hubness}.
	
	
	\section{Conclusions}
	
	We show for the first time that the curse of dimensionality can be broken using Isolation Kernel (IK). It is possible because (a) IK measures the similarity between two points based on the space partitionings produced from an isolation mechanism;
	(b)  the probability of a point falling into a partition in the isolation partitioning of Voronoi diagram is independent of data distribution, the distance measure used to create the Voronoi partitions and the data dimensions, implied in Lemma \ref{lem2}; and (c) IK's unique feature map has its dimensionality linked to a concatenation of these partitionings.
	Theorem~\ref{thm2} suggests that increasing the number of partitionings $t$ (i.e., the dimensionality of the feature map) leads to increased distinguishability, independent of the number of dimensions and distribution in data space.
	
	
	
	Isolation Kernel, with its feature map having the distinguishability stated in Theorem \ref{thm2}, is the key to  consistently producing (i) faster indexed search than brute force search, and high retrieval precision, (ii) good clustering outcomes with SC and DP,  (iii) high accuracy of SVM classification, and (iv) matching structures in t-SNE visualization in both low and high dimensional data spaces. 
	
	Euclidean distance, Gaussian and linear kernels  have a mixed bag of poor and good results in high dimensions in our experiments. This is not a surprising outcome of the curse of dimensionality, echoing the current state of understanding of these measures, with or without the aid of intrinsic dimensions.
	

	
	
	\newpage
	\appendix
	
	\setcounter{Alem}{1}
	\section{Proofs of Lemmas and Theorems}
	\label{sec_A_Proofs}
	\begin{Alem}\label{Alem1}
		Given two distinct points ${\bf x}_a$, ${\bf x}_b$ i.i.d. drawn from $F$ on $\Omega$ 
		and a data set $D = \{{\bf x}_i \in \mathbb{R}^d \; | \; i=1, \dots, n\} \sim G^n$ 
		where every ${\bf x}_i$ is i.i.d drawn from any probability distribution $G$ on 
		$\Omega$, let the feature vectors of ${\bf x}_a$ and ${\bf x}_b$ be 
		$\phi({\bf x}_a) = [\kappa({\bf x}_a,{\bf x}_1),\dots,\kappa({\bf x}_a,{\bf x}_n)]^\top$ and 
		$\phi({\bf x}_b) = [\kappa({\bf x}_b,{\bf x}_1),\dots,\kappa({\bf x}_b,{\bf x}_n)]^\top$ in a feature space $\mathcal{H}_\kappa$.  The following inequality holds:
		\begin{equation}
			P\left(\ell_p(\phi({\bf x}_a)-\phi({\bf x}_b)) \leq 2L\epsilon n^{1/p} \right) \geq (1-2\alpha(\epsilon))^{2n}, \nonumber
		\end{equation}
		for any $\epsilon > 0$, where $\ell_p$ is an $\ell_p$-norm on $\mathcal{H}_\kappa$, and 
		$L$ is the Lipschitz constant of $f_\kappa$.\\
		\\
		Proof. Since $f_\kappa$ is continuous and monotonically decreasing for $m({\bf x},{\bf y})$, the following holds for every ${\bf x}_i$: 
		\[
		A_\epsilon({\bf x}_i) = \{{\bf x} \in \Omega \; | \; f_\kappa(m({\bf x},{\bf x}_i)) \in [f_\kappa(M({\bf x}_i)+\epsilon), f_\kappa(M({\bf x}_i)-\epsilon)]\}
		\]
		\[ = \{{\bf x} \in \Omega \; | \; m({\bf x},{\bf x}_i) \in [M({\bf x}_i)-\epsilon, M({\bf x}_i)+\epsilon]\} \hspace{6mm}
		\]
		\noindent
		Thus, $F(A_\epsilon({\bf x}_i)) \geq 1- 2\alpha(\epsilon)$ holds for every ${\bf x}_i$ 
		from Corollary~\ref{col1}. 
		
		Further, 
		$P({\bf x}_a$, ${\bf x}_b \in A_\epsilon({\bf x}_i))=F(A_\epsilon({\bf x}_i))^2$ holds 
		for every ${\bf x}_i$, since ${\bf x}_a$ and ${\bf x}_b$ are i.i.d. drawn from $F$, and 
		${\bf x}_i$ is i.i.d.
		drawn from $G$. Therefore, the following holds:
		\begin{eqnarray}
			P({\bf x}_a, {\bf x}_b \in A_\epsilon({\bf x}_i) \mbox{ for all } i=1,\dots,n)  & = &\Pi_{i=1}^n P({\bf x}_a, {\bf x}_b \in A_\epsilon({\bf x}_i))\nonumber\\ &  = &\Pi_{i=1}^n F(A_\epsilon({\bf x}_i))^2 \nonumber\\ & \geq &  (1-2\alpha(\epsilon))^{2n} \nonumber
		\end{eqnarray}
		Accordingly, the following holds: 
		\begin{eqnarray}
			P\left(\ell_p(\phi({\bf x}_a)-\phi({\bf x}_b))  \leq
			\left(\sum_{i=1}^n |f_\kappa(M({\bf x}_i)-\epsilon)-f_\kappa(M({\bf x}_i)+\epsilon)|^p \right)^{1/p}\right) \nonumber\\   \geq (1-2\alpha(\epsilon))^{2n}\nonumber
		\end{eqnarray}
		Moreover, the following inequality holds from the Lipschitz continuity of $f_\kappa$.
		\begin{equation}
			|f_\kappa(M({\bf x}_i)-\epsilon)-f_\kappa(M({\bf x}_i)+\epsilon)| \leq L|(M({\bf x}_i)+\epsilon) - (M({\bf x}_i)-\epsilon)| = 2L\epsilon\nonumber
		\end{equation}
		The last two inequalities derive Lemma \ref{lem1}. \qed
	\end{Alem}
	\begin{Athm}\label{Athm1}
		Under the same condition as Lemma~\ref{Alem1}, the feature space $\mathcal{H}_\kappa$ 
		has indistinguishability in the limit of $d \rightarrow \infty$.\\
		\\
		Proof. By choosing $d$ as $d \geq \beta/\epsilon^2$ ($\beta >0$), we obtain the 
		following inequalities from Lemma~\ref{Alem1}.
		\begin{eqnarray}
			&&\alpha(\epsilon) = C_1 e^{-C_2 \epsilon^2 d} \leq  C_1 e^{-C_2 \beta}, \mbox{ and}\nonumber\\
			&&P\left(\ell_p(\phi({\bf x}_a)-\phi({\bf x}_b)) \leq 2L\epsilon n^{1/p}\right) \geq (1-2\alpha(\epsilon))^{2n} \geq \delta,\nonumber
		\end{eqnarray}
		where $\delta = (1-2C_1 e^{-C_2 \beta})^{2n} \in (0,1)$. 
		
		\noindent
		In the limit of $\epsilon \rightarrow 0$, 
		{\it i.e.}, $d \rightarrow \infty$, 
		\begin{equation}
			P(\ell_p(\phi({\bf x}_a)-\phi({\bf x}_b)) = 0) = P(\phi({\bf x}_a) = \phi({\bf x}_b)) \geq \delta.\nonumber
		\end{equation}
		Since this holds for any two distinct points ${\bf x}_a,{\bf x}_b \in \Omega$ and \linebreak $\phi({\bf x}_a), \phi({\bf x}_b) \in \mathcal{H}_\kappa$,
		$\mathcal{H}_\kappa$ has indistinguishability based on Eq \ref{indist}. \qed
	\end{Athm}
	
	\begin{Alem}\label{Alem2}
		Given ${\bf x} \in \Omega$ and a data set 
		$\mathcal D = \{ {\bf z}_1, \dots, {\bf z}_\psi \} \sim G^\psi$ where 
		every ${\bf z}_j$ is i.i.d. drawn from any probability distribution $G$ on 
		$\Omega$ and forms its Voronoi partition $\theta_j \subset \Omega$, the probability that ${\bf z}_j$ is the nearest neighbor 
		of ${\bf x}$ in $\mathcal D$ is given as: $P({\bf x} \in \theta_j) = 1/\psi$, for every 
		$j=1,\dots,\psi$. \\
		\\
		Proof. Let $R({\bf x})=\{{\bf y} \in \Omega \; | \; m({\bf x},{\bf y}) \leq r\}$, 
		$\Delta R({\bf x})=\{{\bf y} \in \Omega \; | \; r \leq m({\bf x},{\bf y}) \leq r+ \Delta r\}$ 
		for $\Delta r > 0$, $u = \int_{R({\bf x})} \rho_G({\bf y})d{\bf y}$ and 
		$\Delta u = \int_{\Delta R({\bf x})} \rho_G({\bf y})d{\bf y}$ where $\rho_G$ is the probability 
		density of $G$. Further, let two events $S$ and $T({\bf z}_j)$ be as follows.
		\begin{eqnarray}
			S &\equiv& \mbox{${\bf z}_k \notin R({\bf x})$ for all ${\bf z}_k \in \mathcal D$ ($k=1,\dots,\psi$)}, \mbox{ and}\nonumber\\ 
			T({\bf z}_j) &\equiv& \left\{ \begin{array}{l} 
				\mbox{${\bf z}_j \in \Delta R({\bf x})$ for ${\bf z}_j \in \mathcal D$},\mbox{ and}\\
				\mbox{${\bf z}_k \notin \Delta R({\bf x})$ for all ${\bf z}_k \in \mathcal D \; (k=1,\dots,\psi, k \neq j)$.}
			\end{array} \right. \nonumber
		\end{eqnarray}
		Then, the probability, that ${\bf z}_j$ is in $\Delta R({\bf x})$ and 
		is the nearest neighbor of ${\bf x}$ in $\mathcal D$, {\it i.e.}, ${\bf x} \in \theta_j$, is 
		$P(S \land T({\bf z}_j))=P(S)P(T({\bf z}_j)|S)$ where 
		\begin{eqnarray}
			P(S) &=& (1-u)^\psi, \mbox{ and}\nonumber\\
			P(T({\bf z}_j)|S) &= & \frac{\Delta u}{1-u} \left\{1-\frac{\Delta u}{1-u} \right\}^{\psi-1}.\nonumber
		\end{eqnarray}
		By letting $\Delta u$ be infinite decimal $du$, $\Delta u/(1-u) \rightarrow du/(1-u)$ 
		and $\{1-\Delta u/(1-u)\} \rightarrow 1$. Then, we obtain the following 
		total probability that ${\bf z}_j$ is the nearest neighbor of ${\bf x}$ in $\mathcal D$, {\it i.e.}, ${\bf x} \in \theta_j$, by integrating $P(S \land T({\bf z}_j))$ on $u \in [0,1]$ for every $\jmath=1,\dots,\psi$.
		\[
		P({\bf x} \in \theta_j) = \int_0^1 (1-u)^\psi \cfrac{du}{1-u} = \cfrac{1}{\psi}\ .
		\] \qed
	\end{Alem}
	This lemma points to a simple but nontrivial result that
	
	\noindent
	$P({\bf x} \in \theta_j)$ 
	is independent of $G$, $m({\bf x},{\bf y})$ and data dimension $d$. 
	
	\begin{Athm}\label{Athm2}
		Given two distinct points ${\bf x}_a$, ${\bf x}_b$ i.i.d. drawn from $F$ on $\Omega$ 
		and ${\mathcal D}_i = \{ {\bf z}_1, \dots, {\bf z}_\psi \} \sim G^\psi$ defined in 
		Lemma~\ref{Alem2}, let the feature vectors of ${\bf x}_a$ and ${\bf x}_b$ be $\Phi({\bf x}_a)$ and $\Phi({\bf x}_b)$
		in a feature space $\mathcal{H}_K$ associated with Isolation Kernel $K$, implemented using Voronoi diagram, as given in Proposition \ref{prop:featureMap}. 
		Then the following holds: 
		\[
		P(\Phi({\bf x}_a)=\Phi({\bf x}_b)) \leq \cfrac{1}{\psi^t},
		\]
		and $\mathcal{H}_K$ 
		always has strong distinguishability for large $\psi$ and $t$.\\
		\\
		Proof. Let $\mathds{1}(\cdot) \in \{0,1\}$ be an indicator function. The following 
		holds for every $\theta_j \in H$; and $H$ is derived from $\mathcal D$ ($j=1,\dots,\psi$) used to implement Isolation Kernel, where $\rho_F$ and $\rho_{G^\psi}$ are 
		the probability density of $F$ and $G^\psi$, respectively. 
		\begin{eqnarray}
			&&\hspace*{-11mm}P({\bf x}_a, {\bf x}_b \in \theta_j)\nonumber\\
			&&\hspace*{-11mm}= \int \int \int \mathds{1}({\bf x}_a, {\bf x}_b \in \theta_j) \rho_{G^\psi}(\mathcal D) \rho_F({\bf x}_a) \rho_F({\bf x}_b) d{\mathcal D} d{\bf x}_a d{\bf x}_b\nonumber
		\end{eqnarray}
		\vspace{-2mm}
		{\small
			\begin{eqnarray}
				&&\hspace*{-7mm}= \int \int \left(\int \mathds{1}({\bf x}_a \in \theta_j) \mathds{1}({\bf x}_b \in \theta_j) \rho_{G^\psi}(\mathcal D) d{\mathcal D}\right) \rho_F({\bf x}_a) \rho_F({\bf x}_b) d{\bf x}_a d{\bf x}_b\nonumber\\
				&&\hspace*{-7mm} \leq \int \int \left( \sqrt{\int \mathds{1}({\bf x}_a \in \theta_j) \rho_{G^\psi}(\mathcal D) d{\mathcal D}} \sqrt{\int \mathds{1}({\bf x}_b \in \theta_j) \rho_{G^\psi}(\mathcal D) d{\mathcal D}} \right) \nonumber\\
				&&\hspace*{-4mm} \times \rho_F({\bf x}_a) \rho_F({\bf x}_b) d{\bf x}_a d{\bf x}_b\nonumber\\
				&&\hspace*{-7mm} = \int \int \sqrt{P({\bf x}_a \in \theta_j|G^\psi)} \sqrt{P({\bf x}_b \in \theta_j|G^\psi)} \rho_F({\bf x}_a) \rho_F({\bf x}_b) d{\bf x}_a d{\bf x}_b\nonumber
		\end{eqnarray}}
		\vspace{-2mm}
		\begin{equation}
			\hspace*{-37mm} = \int \int \frac{1}{\psi} \rho_F({\bf x}_a) \rho_F({\bf x}_b) d{\bf x}_a d{\bf x}_b = \frac{1}{\psi},\nonumber
		\end{equation}
		where the inequality is from the Cauchy-Schwarz inequality 
		and the last line is from Lemma~\ref{Alem2}. Accordingly, the following 
		holds, because ${\mathcal D}_i$ is i.i.d. drawn from $G^\psi$ over $i = 1, \dots, t$.
		\begin{equation}
			P(\Phi({\bf x}_a)=\Phi({\bf x}_b)) = \prod_{i=1}^tP({\bf x}_a, {\bf x}_b \in \theta_j) \leq \frac{1}{\psi^t}\nonumber
		\end{equation}
		\noindent
		where $\theta_j \in H_i$; and $H_i$ is derived from $\mathcal{D}_i$. 
		
		Let $\delta = 1/\psi^t$,  Eq \ref{eqn_distinguishability} holds for any two distinct points ${\bf x}_a,{\bf x}_b \in \Omega$. 
		Thus $\mathcal{H}_K$ 
		has distinguishability. Particularly, $\mathcal{H}_K$ has strong distinguishability if $\psi$ and $t$ are large where $1/\psi^t \ll 1$ holds.\qed
	\end{Athm}
	
	\section{Experimental settings}
	\label{ap1}
	
	We implement Isolation Kernel with the Matlab R2021a. The parameter $t=200$ is used for all the experiments using Isolation Kernel (IK).
	
	
	For the experiments on instability of a measure (reported in Section \ref{sec_instability}), the similarity scores are normalized to [0,1] for all measures.
	
	In the ball tree search experiments, IK feature map is used to map each point in the given dataset into a point in Hilbert space; and for linear kernel (LK), each point is normalized with the square root of its self similarity using LK.  Then exactly the same ball tree index using the Euclidean distance is employed to perform the indexing for LK, IK and distance. The parameter search ranges for IK are $\psi\in \{3,5,..,250\}$.  
	
	We used the package ``sklearn.neighbors'' from Python to conduct the indexing task. The leaf size is 15 and we query the 5 nearest neighbors of each point from the given dataset using the brute-force and ball tree indexing.
	The dataset sizes have been limited to 10,000 or less because of the memory requirements of the indexing algorithm.
	We have also omitted the comparison with AG and SNN \citep{SNN} because they require k-nearest neighbor (kNN) search. It does not make sense to perform indexing using AG/SNN to speed up an NN search because AG/SNN requires a kNN search (which itself requires an index to speed up the search.)
	
	Table \ref{tbl_SC_parameters} shows the search ranges of the parameters for the two clustering algorithms and kernels. All datasets are normalised using the $min$-$max$ normalisation to yield each attribute to be in [0,1] before the clustering begins.
	
	\begin{table}[h]
		\renewcommand{\arraystretch}{1.2}
		\setlength{\tabcolsep}{1.1pt}
		\centering
		\caption{Parameters and their search ranges in spectral clustering (SC) and density-peaks (DP) clustering. $m$ is the maximum pairwise distance in the dataset.}
		\begin{tabular}{|c|c|}
			\hline
			Algorithm & Parameters and their search range \\
			\hline 
			IK & $t=200$; $\psi \in \{2^1, 2^2,..., 2^{10}\}$ \\
			GK &  $\sigma= d \times \mu; \mu \in \{ 2^{-5}, 2^{-4}, ..., 2^{5}\}$ \\
			Adaptive GK & $k \in \{0.05n, 0.1n, 0.15n, ..., 0.45n, 0.5n\}$ \\
			\hdashline 
			DP & $k= \# Cluster$, $\epsilon \in \{ 1\%m,2\%m,...,99\%m \}$\\
			SC  &   $k= \# Cluster$ \\
			\hline
		\end{tabular}%
		\label{tbl_SC_parameters}%
	\end{table}%

	In Section \ref{sec_traditional_tasks}, SVM classifiers in scikit-learn \citep{scikit-learn} are used with IK/LK and Gaussian kernel. 
	They are based on LIBLINEAR \citep{fan2008liblinear} and LIBSVM \citep{CC01a}.  
	
	Table \ref{tbl_SVM_parameters} shows the search ranges of the kernel parameters for SVM classifier. A 5-fold cross validation on the training set is used to determine the best parameter. The reported accuracy in Table~\ref{tbl:SVM-accuracy} is the accuracy obtained from the test set after the final model is trained using the training set and the 5-fold CV determined best parameter.
	\begin{table}[h]
		\centering
		\caption{Kernel parameters and their search ranges in SVM.}
		\begin{tabular}{|c|c|}
			\hline
			& Parameters and their search range \\
			\hline 
			IK & $\psi \in \{ 2^m\ |\ m=2,3,\dots,12\}$ \\   
			GK &  $\sigma= d \times \mu; \mu \in \{ 2^{-5}, 2^{-4}, ..., 2^{5}\}$ (dense datasets)\\
			& $\sigma= \mu; \mu \in \{ 2^{-5}, 2^{-4}, ..., 2^{5}\}$ (sparse datasets: $nnz\% < 1$) \\
			\hline
		\end{tabular}%
		\label{tbl_SVM_parameters}%
	\end{table}%
	

	For the t-SNE \citep{Hinton2008Visualizing} experiment, we set  $tolerance =0.00005$. 
	We report the best visualised results from a search in $[5, 20, 40, 60, 80, 100, \\ 250, 500, 800]$ for both $\psi$ (using IK) and $perplexity$ (using GK). When using IK in t-SNE visualization, we replace the similarity matrix calculated by GK with the similarity matrix calculated by IK. For both FIt-SNE \citep{linderman2019fast} and q-SNE \citep{hakkinen2020qsne}, we report  the best visualization result from the same $perplexity$ search range and fix all other parameters to the default values. 
	
	

	All datasets used are obtained from \url{https://www.csie.ntu.edu.tw/~cjlin/libsvmtools/datasets/}, except Gaussians and $w$-Gaussians which are our creation. 
	

	The machine used in the experiments has one Intel E5-2682 v4 @ 2.50GHz 16 cores CPU with 256GB memory.
	
	\subsection{A guide for parameter setting of IK}
	Some advice on parameter setting is in order when using IK in practice.
	First, finding the `right' parameters for IK may not be an easy task for some applications such as clustering (so as indexing.) This is a general problem in the area of unsupervised learning, not specific to the use of IK. For any kernel employed, when no or insufficient labels are available in a given dataset, it is unclear how an appropriate setting can be found in practice.
	
	Second, $t$ can often be set as default to 200 initially. Then, search for the `right' $\psi$ for a dataset. This is equivalent to searching the bandwidth parameter for Gaussian kernel. Once the $\psi$ setting has been determined, one may increase $t$ to examine whether high accuracy can be achieved (not attempted in our experiments.)
	
	\subsection{IK feature mapping time and SVM runtime}
	\label{sec_IK_mapping_time}
	Table \ref{runt} presents the separate runtimes of IK feature mapping and SVM used to complete the experiments in SVM classification.
	
	\begin{table}[h]
		\centering
		\caption{Runtimes of IK feature mapping and SVM in CPU seconds.}
		\begin{tabular}{|c|r|rr|}
			\hline
			Dataset  & $\psi$   & Mapping & SVM \\
			\hline
			Url   & 32    & 2 & 1\\
			News20  & 64    & 41 & 19\\
			Rcv1  & 64    & 394 & 26\\
			Real-sim  & 128   & 44 & 13\\
			Gaussians & 4     & 8 & 0.5\\
			$w$-Gaussians   & 4  & 7 & 0.5 \\
			Cifar-10 & 128   & 594 & 493\\
			Mnist & 64    & 31 & 17\\
			A9a   & 64    & 9 & 22\\
			Ijcnn1 & 128   & 26 & 40\\
			\hline
		\end{tabular}%
		\label{runt}
	\end{table}

	\section{Estimated intrinsic dimensions using TLE}
	\label{sec_ID}
	
\begin{figure}[h]
		\centering
		\includegraphics[width=3.5in]{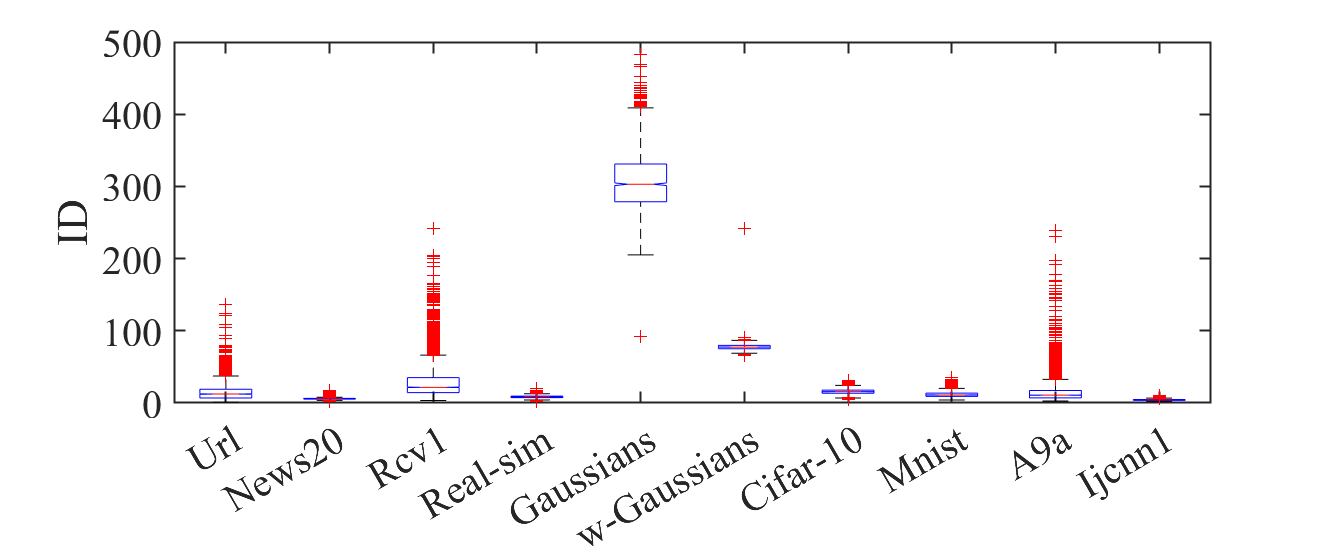}
		\caption{Estimated intrinsic dimensions (ID) for 10 datasets using TLE with $k=50$. }
		\label{fig_ID}
	\end{figure}

	Figure \ref{fig_ID} shows that intrinsic dimensions (ID) as estimated by a recent local ID estimator TLE \citep{LID-SIAM19}. 
	
	It is interesting to note that TLE has significantly underestimated the IDs of the Gaussians and $w$-Gaussians datasets which have true 10,000 and 5,000 IDs, respectively. This is likely to be due to the estimator's inability to deal with high dimensional IDs. Most papers used a low dimensional Gaussian distribution as ground truth in their experiments, e.g., \citep{LID-SIAM19}.

	With these under-estimations, one may use the typical high IDs to explain the low accuracy of SVM using Gaussian kernel in the $w$-Gaussians dataset (as TLE estimated it to be more than 75 IDs). But it could not explain SVM's high accuracy in the Gaussians dataset (as it was estimated to be more than 300 IDs.)  
	
	\section{The influence of partitionings on distinguishability}
	\label{sec_partitiongs}
	This section investigates two influences of partitionings on $N_\epsilon$ used in Section \ref{sec_instability}: (i) the number of partitionings $t$; and (ii) the partitions are generated using a dataset different from the given dataset.
	
\begin{figure}[h]
		\centering
		\begin{subfigure}[b]{0.234\textwidth}
			\centering
			\includegraphics[width=1.5in]{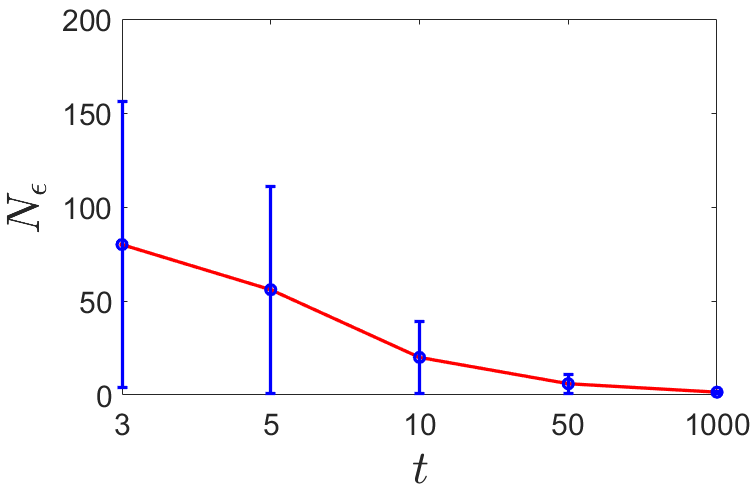}
			\caption{IK (given dataset)}
		\end{subfigure}
		\begin{subfigure}[b]{0.234\textwidth}
			\centering
			\includegraphics[width=1.5in]{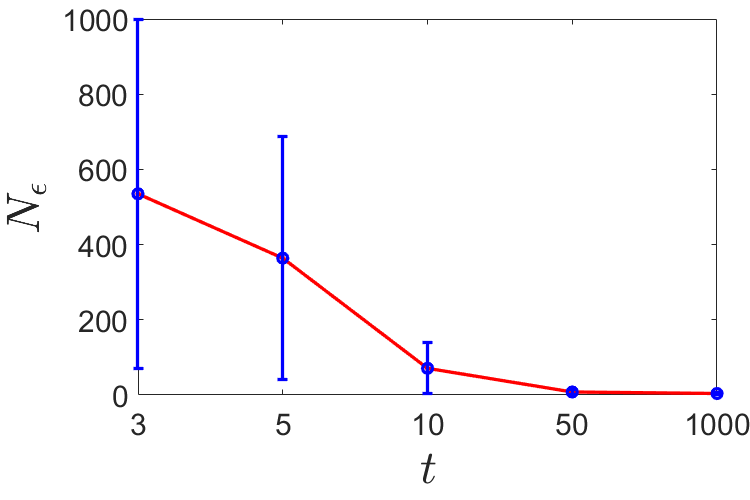}
			\caption{IK (uniform data distribution)}
		\end{subfigure} 
		\caption{$N_\epsilon$ ($\epsilon=0.005$) as a result of varying $t$ on $d=10000$. The same dataset in Figure 1(a) is used.  The result for each $t$ value is the average and standard error over 10 trials. } 
		\label{fig_Varyingt} 
	\end{figure}

	Figure \ref{fig_Varyingt}(a) shows that $N_\epsilon$ decreases as $t$ increases.  This effect is exactly as predicted in Theorem \ref{thm2}, i.e., increasing $t$ leads to increased distinguishability.

	It is possible to derive an IK using a dataset of uniform distribution (which is different from the given dataset.) Figure \ref{fig_Varyingt}(b) shows the outcome, i.e., it exhibits the same phenomenon as the IK derived from the given dataset, except that small $t$ leads to poorer distinguishability (having higher $N_\epsilon$ and higher variance.) 
	
	At high $t$, there is virtually no difference between the two versions of IK. This is the direct outcome of Lemma \ref{lem2}: the probability of any point in the data space falling into one of the $\psi$ partitions is independent of the dataset used to generate the partitions.

	\section{Hubness effect}
	\label{App_hubness}
	
	\citet{HubsInSpace-JMLR-2010} attribute the hubness effect to a consequence of high (intrinsic) dimensionality of data, and not factors such as sparsity and skewness of the distribution. However, it is unclear why hubness only occurs in $k$-nearest neighborhood, and not in $\epsilon$-neighborhood on the same dataset.
	
	In the context of $k$ nearest neighborhood, it has been shown that there are few points which are the nearest neighbors to many points in a dataset of high dimensions \citep{HubsInSpace-JMLR-2010}.
	Let $N_k(y)$ be the set of $k$ nearest neighbors of $y$; and $k$-occurrences of $x$,
	$O_k (x) = |\{y : x \in N_k (y)\}|$, be the number of other points in the given dataset where $x$ is one of their $k$ nearest neighbors.
	As the number of dimensions increases, the distribution of $O_k(x)$ becomes considerably skewed (i.e., there are many points with zero or small $O_k$ and only a few points have large $O_k(\cdot)$ for many widely used distance measures \citep{HubsInSpace-JMLR-2010}. The points with large $O_k(\cdot)$ are considered as `hubs', i.e., the popular nearest neighbors.
	
\begin{figure}[h]
		\centering
		\begin{subfigure}[b]{0.15\textwidth}
			\centering
			\includegraphics[width=1.15in]{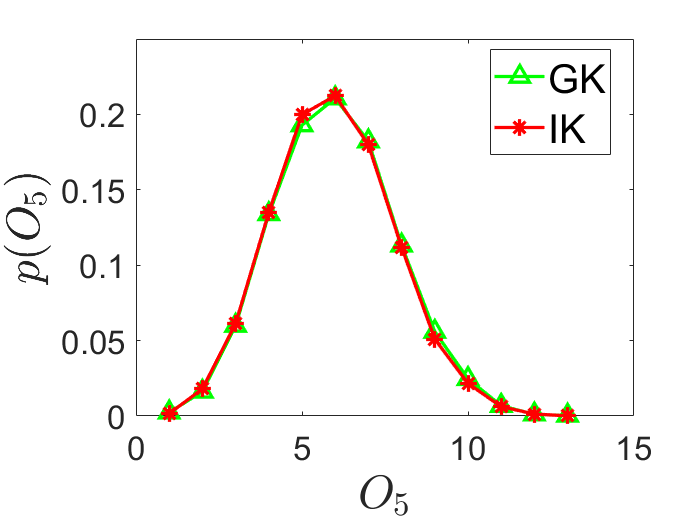}
			\caption{$d=3$}
		\end{subfigure} 
		\begin{subfigure}[b]{0.15\textwidth}
			\centering
			\includegraphics[width=1.15in]{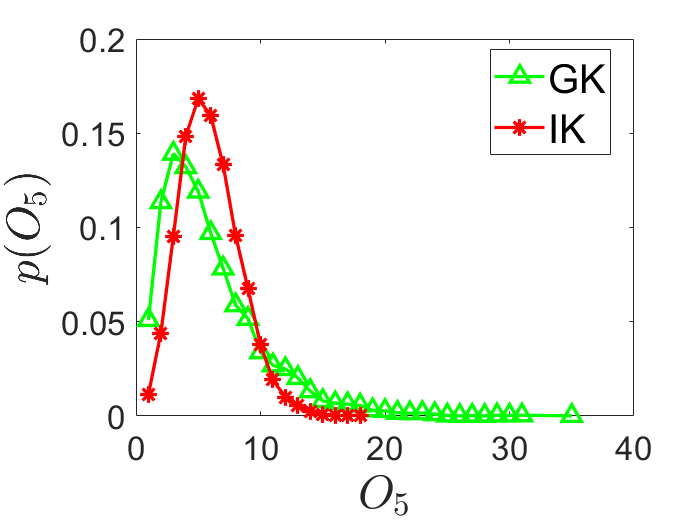}
			\caption{$d=20$ }
		\end{subfigure}
		\begin{subfigure}[b]{0.15\textwidth}
			\centering
			\includegraphics[width=1.15in]{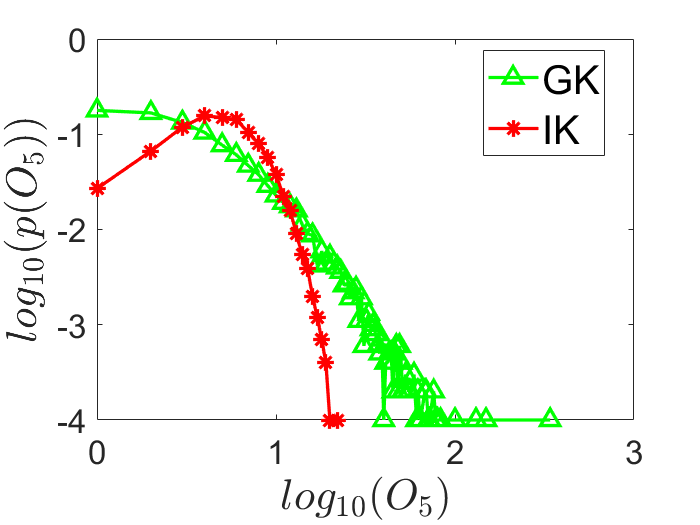}
			\caption{$d=100$ }
		\end{subfigure}
		\caption{The effect of hubness in $k$-nearest neighbors: GK vs IK. The experiment setting is as used by \cite{HubsInSpace-JMLR-2010}: a random dataset is drawn uniformly from the unit hypercube $[0,1]^d$. The parameter settings used are: $\psi=32$ for IK;
			$\sigma=5$ for GK.}
		\label{fig_hubness} 
	\end{figure}

	Figure \ref{fig_hubness} shows the result of $O_5$ vs $p(O_5)$ comparing Gaussian kernel and IK, where $p(O_k(x)) = \frac{|\{y \in D | O_k(y)=O_k(x) \}|}{|D|}$.
	Consistent with the result shown by \citet{HubsInSpace-JMLR-2010} for distance measure, this result shows that Gaussian kernel suffers from the hubness effect as the number of dimensions increases. In contrast, IK is not severely affected by the hubness effect.
	
	\newpage
	\bibliographystyle{ACM-Reference-Format}
	\bibliography{references}


\begin{thebibliography}{50}


\ifx \showCODEN    \undefined \def \showCODEN     #1{\unskip}     \fi
\ifx \showDOI      \undefined \def \showDOI       #1{#1}\fi
\ifx \showISBNx    \undefined \def \showISBNx     #1{\unskip}     \fi
\ifx \showISBNxiii \undefined \def \showISBNxiii  #1{\unskip}     \fi
\ifx \showISSN     \undefined \def \showISSN      #1{\unskip}     \fi
\ifx \showLCCN     \undefined \def \showLCCN      #1{\unskip}     \fi
\ifx \shownote     \undefined \def \shownote      #1{#1}          \fi
\ifx \showarticletitle \undefined \def \showarticletitle #1{#1}   \fi
\ifx \showURL      \undefined \def \showURL       {\relax}        \fi
\providecommand\bibfield[2]{#2}
\providecommand\bibinfo[2]{#2}
\providecommand\natexlab[1]{#1}
\providecommand\showeprint[2][]{arXiv:#2}

\bibitem[\protect\citeauthoryear{Aggarwal, Hinneburg, and Keim}{Aggarwal
  et~al\mbox{.}}{2001}]%
        {Aggarwal-2001}
\bibfield{author}{\bibinfo{person}{Charu~C. Aggarwal},
  \bibinfo{person}{Alexander Hinneburg}, {and} \bibinfo{person}{Daniel~A.
  Keim}.} \bibinfo{year}{2001}\natexlab{}.
\newblock \showarticletitle{On the surprising behavior of distance metrics in
  high dimensional space}. In \bibinfo{booktitle}{\emph{Proceedings of the 8th
  International Conference on Database Theory}}. \bibinfo{pages}{420--434}.
\newblock


\bibitem[\protect\citeauthoryear{Amsaleg, Chelly, Houle, Kawarabayashi,
  Radovanovic, and Treeratanajaru}{Amsaleg et~al\mbox{.}}{2019}]%
        {LID-SIAM19}
\bibfield{author}{\bibinfo{person}{Laurent Amsaleg}, \bibinfo{person}{Oussama
  Chelly}, \bibinfo{person}{Michael~E. Houle}, \bibinfo{person}{Ken{-}ichi
  Kawarabayashi}, \bibinfo{person}{Milos Radovanovic}, {and}
  \bibinfo{person}{Weeris Treeratanajaru}.} \bibinfo{year}{2019}\natexlab{}.
\newblock \showarticletitle{Intrinsic Dimensionality Estimation within Tight
  Localities}. In \bibinfo{booktitle}{\emph{Proceedings of the 2019 {SIAM}
  International Conference on Data Mining}}. \bibinfo{pages}{181--189}.
\newblock


\bibitem[\protect\citeauthoryear{Aryal, Ting, Washio, and Haffari}{Aryal
  et~al\mbox{.}}{2017}]%
        {Aryal2017}
\bibfield{author}{\bibinfo{person}{Sunil Aryal}, \bibinfo{person}{Kai~Ming
  Ting}, \bibinfo{person}{Takashi Washio}, {and} \bibinfo{person}{Gholamreza
  Haffari}.} \bibinfo{year}{2017}\natexlab{}.
\newblock \showarticletitle{Data-dependent dissimilarity measure: an effective
  alternative to geometric distance measures}.
\newblock \bibinfo{journal}{\emph{Knowledge and Information Systems}}
  \bibinfo{volume}{53}, \bibinfo{number}{2} (\bibinfo{date}{01 Nov}
  \bibinfo{year}{2017}), \bibinfo{pages}{479--506}.
\newblock


\bibitem[\protect\citeauthoryear{Bennett, Fayyad, and Geiger}{Bennett
  et~al\mbox{.}}{1999}]%
        {Density-based-Indexing-KDD1999}
\bibfield{author}{\bibinfo{person}{Kristin~P. Bennett}, \bibinfo{person}{Usama
  Fayyad}, {and} \bibinfo{person}{Dan Geiger}.}
  \bibinfo{year}{1999}\natexlab{}.
\newblock \showarticletitle{Density-Based Indexing for Approximate
  Nearest-Neighbor Queries}. In \bibinfo{booktitle}{\emph{Proceedings of the
  Fifth ACM SIGKDD International Conference on Knowledge Discovery and Data
  Mining}}. \bibinfo{pages}{233--243}.
\newblock


\bibitem[\protect\citeauthoryear{Beyer, Goldstein, Ramakrishnan, and
  Shaft}{Beyer et~al\mbox{.}}{1999}]%
        {Beyer:1999}
\bibfield{author}{\bibinfo{person}{Kevin~S. Beyer}, \bibinfo{person}{Jonathan
  Goldstein}, \bibinfo{person}{Raghu Ramakrishnan}, {and} \bibinfo{person}{Uri
  Shaft}.} \bibinfo{year}{1999}\natexlab{}.
\newblock \showarticletitle{When Is ``Nearest Neighbor'' Meaningful?}. In
  \bibinfo{booktitle}{\emph{Proceedings of the 7th International Conference on
  Database Theory}}. \bibinfo{publisher}{Springer-Verlag},
  \bibinfo{address}{London, UK}, \bibinfo{pages}{217--235}.
\newblock


\bibitem[\protect\citeauthoryear{Chang and Lin}{Chang and Lin}{2011}]%
        {CC01a}
\bibfield{author}{\bibinfo{person}{Chih-Chung Chang} {and}
  \bibinfo{person}{Chih-Jen Lin}.} \bibinfo{year}{2011}\natexlab{}.
\newblock \showarticletitle{{LIBSVM}: A library for support vector machines}.
\newblock \bibinfo{journal}{\emph{ACM Transactions on Intelligent Systems and
  Technology}}  \bibinfo{volume}{2} (\bibinfo{year}{2011}),
  \bibinfo{pages}{27:1--27}.
\newblock
Issue 3.


\bibitem[\protect\citeauthoryear{Chen, Song, Bai, Lin, and Chang}{Chen
  et~al\mbox{.}}{2010}]%
        {chen2010parallel}
\bibfield{author}{\bibinfo{person}{Wen-Yen Chen}, \bibinfo{person}{Yangqiu
  Song}, \bibinfo{person}{Hongjie Bai}, \bibinfo{person}{Chih-Jen Lin}, {and}
  \bibinfo{person}{Edward~Y Chang}.} \bibinfo{year}{2010}\natexlab{}.
\newblock \showarticletitle{Parallel spectral clustering in distributed
  systems}.
\newblock \bibinfo{journal}{\emph{IEEE transactions on pattern analysis and
  machine intelligence}} \bibinfo{volume}{33}, \bibinfo{number}{3}
  (\bibinfo{year}{2010}), \bibinfo{pages}{568--586}.
\newblock


\bibitem[\protect\citeauthoryear{Durrant and Kab\'{a}n}{Durrant and
  Kab\'{a}n}{2009}]%
        {WhenNNisMeaningful-2009}
\bibfield{author}{\bibinfo{person}{Robert~J. Durrant} {and}
  \bibinfo{person}{Ata Kab\'{a}n}.} \bibinfo{year}{2009}\natexlab{}.
\newblock \showarticletitle{When is `nearest Neighbour' Meaningful: A Converse
  Theorem and Implications}.
\newblock \bibinfo{journal}{\emph{Journal of Complexity}} \bibinfo{volume}{25},
  \bibinfo{number}{4} (\bibinfo{year}{2009}), \bibinfo{pages}{385--397}.
\newblock


\bibitem[\protect\citeauthoryear{Ester, Kriegel, Sander, and Xu}{Ester
  et~al\mbox{.}}{1996}]%
        {ester1996density}
\bibfield{author}{\bibinfo{person}{Martin Ester}, \bibinfo{person}{Hans-Peter
  Kriegel}, \bibinfo{person}{J{\"o}rg Sander}, {and} \bibinfo{person}{Xiaowei
  Xu}.} \bibinfo{year}{1996}\natexlab{}.
\newblock \showarticletitle{A density-based algorithm for discovering clusters
  in large spatial databases with noise}. In
  \bibinfo{booktitle}{\emph{Proceedings of the Second International Conference
  on Knowledge Discovery and Data Mining}}. \bibinfo{pages}{226--231}.
\newblock


\bibitem[\protect\citeauthoryear{Fan, Chang, Hsieh, Wang, and Lin}{Fan
  et~al\mbox{.}}{2008}]%
        {fan2008liblinear}
\bibfield{author}{\bibinfo{person}{Rong-En Fan}, \bibinfo{person}{Kai-Wei
  Chang}, \bibinfo{person}{Cho-Jui Hsieh}, \bibinfo{person}{Xiang-Rui Wang},
  {and} \bibinfo{person}{Chih-Jen Lin}.} \bibinfo{year}{2008}\natexlab{}.
\newblock \showarticletitle{LIBLINEAR: A library for large linear
  classification}.
\newblock \bibinfo{journal}{\emph{Journal of Machine Learning Research}}
  (\bibinfo{year}{2008}), \bibinfo{pages}{1871--1874}.
\newblock


\bibitem[\protect\citeauthoryear{Felix, Suresh, Choromanski, Holtmann-Rice, and
  Kumar}{Felix et~al\mbox{.}}{2016}]%
        {OrthonogalRandomFeatures2016}
\bibfield{author}{\bibinfo{person}{X.~Y. Felix}, \bibinfo{person}{A.~T.
  Suresh}, \bibinfo{person}{K.~M. Choromanski}, \bibinfo{person}{D.~N.
  Holtmann-Rice}, {and} \bibinfo{person}{S. Kumar}.}
  \bibinfo{year}{2016}\natexlab{}.
\newblock \showarticletitle{Orthogonal random features}. In
  \bibinfo{booktitle}{\emph{Advances in Neural Information Processing
  Systems}}. \bibinfo{pages}{1975--1983}.
\newblock


\bibitem[\protect\citeauthoryear{Francois, Wertz, and Verleysen}{Francois
  et~al\mbox{.}}{2007}]%
        {Concentration-Fractionaldistances}
\bibfield{author}{\bibinfo{person}{Damien Francois}, \bibinfo{person}{Vincent
  Wertz}, {and} \bibinfo{person}{Michel Verleysen}.}
  \bibinfo{year}{2007}\natexlab{}.
\newblock \showarticletitle{The concentration of fractional distances}.
\newblock \bibinfo{journal}{\emph{IEEE Transactions on Knowledge and Data
  Engineering}} \bibinfo{volume}{19}, \bibinfo{number}{7}
  (\bibinfo{year}{2007}), \bibinfo{pages}{873--886}.
\newblock


\bibitem[\protect\citeauthoryear{Fukunaga}{Fukunaga}{2013}]%
        {Fukunaga}
\bibfield{author}{\bibinfo{person}{Keinosuke Fukunaga}.}
  \bibinfo{year}{2013}\natexlab{}.
\newblock \bibinfo{booktitle}{\emph{Introduction to Statistical Pattern
  Recognition} (\bibinfo{edition}{2nd} ed.)}.
\newblock \bibinfo{publisher}{Academic Press}, Chapter 6, section 6.2.
\newblock


\bibitem[\protect\citeauthoryear{Gromov and Milman}{Gromov and Milman}{1983}]%
        {Gromov}
\bibfield{author}{\bibinfo{person}{Mikhael Gromov} {and}
  \bibinfo{person}{Vitali Milman}.} \bibinfo{year}{1983}\natexlab{}.
\newblock \showarticletitle{A topological application of the isoperimetric
  inequality}.
\newblock \bibinfo{journal}{\emph{American Journal of Mathematics}}
  \bibinfo{volume}{105}, \bibinfo{number}{4} (\bibinfo{year}{1983}),
  \bibinfo{pages}{843--854}.
\newblock


\bibitem[\protect\citeauthoryear{H{\"a}kkinen, Koiranen, Casado, Kaipio,
  Lehtonen, Petrucci, Hynninen, Hietanen, Carp{\'e}n, Pasquini,
  et~al\mbox{.}}{H{\"a}kkinen et~al\mbox{.}}{2020}]%
        {hakkinen2020qsne}
\bibfield{author}{\bibinfo{person}{Antti H{\"a}kkinen}, \bibinfo{person}{Juha
  Koiranen}, \bibinfo{person}{Julia Casado}, \bibinfo{person}{Katja Kaipio},
  \bibinfo{person}{Oskari Lehtonen}, \bibinfo{person}{Eleonora Petrucci},
  \bibinfo{person}{Johanna Hynninen}, \bibinfo{person}{Sakari Hietanen},
  \bibinfo{person}{Olli Carp{\'e}n}, \bibinfo{person}{Luca Pasquini},
  {et~al\mbox{.}}} \bibinfo{year}{2020}\natexlab{}.
\newblock \showarticletitle{qSNE: quadratic rate t-SNE optimizer with automatic
  parameter tuning for large datasets}.
\newblock \bibinfo{journal}{\emph{Bioinformatics}} \bibinfo{volume}{36},
  \bibinfo{number}{20} (\bibinfo{year}{2020}), \bibinfo{pages}{5086--5092}.
\newblock


\bibitem[\protect\citeauthoryear{Hinneburg, Aggarwal, and Keim}{Hinneburg
  et~al\mbox{.}}{2000}]%
        {NN-HighDim-VLDB-2000}
\bibfield{author}{\bibinfo{person}{Alexander Hinneburg},
  \bibinfo{person}{Charu~C. Aggarwal}, {and} \bibinfo{person}{Daniel~A. Keim}.}
  \bibinfo{year}{2000}\natexlab{}.
\newblock \showarticletitle{What Is the Nearest Neighbor in High Dimensional
  Spaces?}. In \bibinfo{booktitle}{\emph{Proceedings of the 26th International
  Conference on Very Large Data Bases}}. \bibinfo{pages}{506--515}.
\newblock


\bibitem[\protect\citeauthoryear{Houle, Kriegel, Kr{\"o}ger, Schubert, and
  Zimek}{Houle et~al\mbox{.}}{2010}]%
        {SNN-Defeat-Curse?}
\bibfield{author}{\bibinfo{person}{Michael~E. Houle},
  \bibinfo{person}{Hans-Peter Kriegel}, \bibinfo{person}{Peer Kr{\"o}ger},
  \bibinfo{person}{Erich Schubert}, {and} \bibinfo{person}{Arthur Zimek}.}
  \bibinfo{year}{2010}\natexlab{}.
\newblock \showarticletitle{Can Shared-Neighbor Distances Defeat the Curse of
  Dimensionality?}. In \bibinfo{booktitle}{\emph{Proceedings of the
  International Conference on Scientific and Statistical Database Management}}.
  \bibinfo{pages}{482--500}.
\newblock


\bibitem[\protect\citeauthoryear{Jarvis and Patrick}{Jarvis and
  Patrick}{1973}]%
        {SNN}
\bibfield{author}{\bibinfo{person}{Raymond~A. Jarvis} {and}
  \bibinfo{person}{Edward~A. Patrick}.} \bibinfo{year}{1973}\natexlab{}.
\newblock \showarticletitle{Clustering using a similarity measure based on
  shared near neighbors}.
\newblock \bibinfo{journal}{\emph{IEEE Trans. Comput.}} \bibinfo{volume}{100},
  \bibinfo{number}{11} (\bibinfo{year}{1973}), \bibinfo{pages}{1025--1034}.
\newblock


\bibitem[\protect\citeauthoryear{Jolliffe}{Jolliffe}{2002}]%
        {PCA-Book2002}
\bibfield{author}{\bibinfo{person}{I.~T. Jolliffe}.}
  \bibinfo{year}{2002}\natexlab{}.
\newblock \bibinfo{booktitle}{\emph{Principal Component Analysis}}.
\newblock \bibinfo{publisher}{Springer Series in Statistics. New York:
  Springer-Verlag}.
\newblock


\bibitem[\protect\citeauthoryear{Kab\'{a}n}{Kab\'{a}n}{2011}]%
        {Ata-Kaban-2011}
\bibfield{author}{\bibinfo{person}{Ata Kab\'{a}n}.}
  \bibinfo{year}{2011}\natexlab{}.
\newblock \showarticletitle{On the Distance Concentration Awareness of Certain
  Data Reduction Techniques}.
\newblock \bibinfo{journal}{\emph{Pattern Recognition}} \bibinfo{volume}{44},
  \bibinfo{number}{2} (\bibinfo{year}{2011}), \bibinfo{pages}{265--277}.
\newblock


\bibitem[\protect\citeauthoryear{Kibriya and Frank}{Kibriya and Frank}{2007}]%
        {ExactNNSearch-PKDD2007}
\bibfield{author}{\bibinfo{person}{Ashraf~M. Kibriya} {and}
  \bibinfo{person}{Eibe Frank}.} \bibinfo{year}{2007}\natexlab{}.
\newblock \showarticletitle{An Empirical Comparison of Exact Nearest Neighbour
  Algorithms}. In \bibinfo{booktitle}{\emph{Proceedings of European Conference
  on Principles of Data Mining and Knowledge Discovery}}.
  \bibinfo{pages}{140--151}.
\newblock


\bibitem[\protect\citeauthoryear{Kpotufe}{Kpotufe}{2011}]%
        {LID-NIPS2011}
\bibfield{author}{\bibinfo{person}{Samory Kpotufe}.}
  \bibinfo{year}{2011}\natexlab{}.
\newblock \showarticletitle{{K-NN} Regression Adapts to Local Intrinsic
  Dimension}. In \bibinfo{booktitle}{\emph{Proceedings of the 24th
  International Conference on Neural Information Processing Systems}}.
  \bibinfo{pages}{729--737}.
\newblock


\bibitem[\protect\citeauthoryear{Kriegel, Schubert, and Zimek}{Kriegel
  et~al\mbox{.}}{2008}]%
        {Angle-BasedAD-KDD2008}
\bibfield{author}{\bibinfo{person}{Hans-Peter Kriegel},
  \bibinfo{person}{Matthias Schubert}, {and} \bibinfo{person}{Arthur Zimek}.}
  \bibinfo{year}{2008}\natexlab{}.
\newblock \showarticletitle{Angle-Based Outlier Detection in High-Dimensional
  Data}. In \bibinfo{booktitle}{\emph{Proceedings of the 14th ACM SIGKDD
  International Conference on Knowledge Discovery and Data Mining}}.
  \bibinfo{publisher}{Association for Computing Machinery},
  \bibinfo{pages}{444--452}.
\newblock


\bibitem[\protect\citeauthoryear{Ledoux}{Ledoux}{2001}]%
        {Ledoux}
\bibfield{author}{\bibinfo{person}{Michel Ledoux}.}
  \bibinfo{year}{2001}\natexlab{}.
\newblock \bibinfo{booktitle}{\emph{The Concentration of Measure Phenomenon}}.
\newblock \bibinfo{publisher}{Mathematical Surveys \& Monographs, The American
  Mathematical Society}.
\newblock


\bibitem[\protect\citeauthoryear{Linderman, Rachh, Hoskins, Steinerberger, and
  Kluger}{Linderman et~al\mbox{.}}{2019}]%
        {linderman2019fast}
\bibfield{author}{\bibinfo{person}{George~C Linderman}, \bibinfo{person}{Manas
  Rachh}, \bibinfo{person}{Jeremy~G Hoskins}, \bibinfo{person}{Stefan
  Steinerberger}, {and} \bibinfo{person}{Yuval Kluger}.}
  \bibinfo{year}{2019}\natexlab{}.
\newblock \showarticletitle{Fast interpolation-based t-SNE for improved
  visualization of single-cell RNA-seq data}.
\newblock \bibinfo{journal}{\emph{Nature methods}} \bibinfo{volume}{16},
  \bibinfo{number}{3} (\bibinfo{year}{2019}), \bibinfo{pages}{243--245}.
\newblock


\bibitem[\protect\citeauthoryear{Liu, Ting, and Zhou}{Liu
  et~al\mbox{.}}{2008}]%
        {liu2008isolation}
\bibfield{author}{\bibinfo{person}{Fei~Tony Liu}, \bibinfo{person}{Kai~Ming
  Ting}, {and} \bibinfo{person}{Zhi-Hua Zhou}.}
  \bibinfo{year}{2008}\natexlab{}.
\newblock \showarticletitle{Isolation forest}. In
  \bibinfo{booktitle}{\emph{Proceedings of the IEEE International Conference on
  Data Mining}}. \bibinfo{pages}{413--422}.
\newblock


\bibitem[\protect\citeauthoryear{Meister and Steinwart}{Meister and
  Steinwart}{2016}]%
        {VP-SVM-2016}
\bibfield{author}{\bibinfo{person}{Mona Meister} {and} \bibinfo{person}{Ingo
  Steinwart}.} \bibinfo{year}{2016}\natexlab{}.
\newblock \showarticletitle{Optimal Learning Rates for Localized SVMs}.
\newblock \bibinfo{journal}{\emph{Journal of Machine Learning Research}}
  \bibinfo{volume}{17}, \bibinfo{number}{1} (\bibinfo{year}{2016}),
  \bibinfo{pages}{6722–6765}.
\newblock


\bibitem[\protect\citeauthoryear{Mil{\textquotesingle}man}{Mil{\textquotesingle}man}{1972}]%
        {Mil_man_1972}
\bibfield{author}{\bibinfo{person}{Vitali~D. Mil{\textquotesingle}man}.}
  \bibinfo{year}{1972}\natexlab{}.
\newblock \showarticletitle{New proof of the theorem of {A}. {D}voretzky on
  intersections of convex bodies}.
\newblock \bibinfo{journal}{\emph{Functional Analysis and Its Applications}}
  \bibinfo{volume}{5}, \bibinfo{number}{4} (\bibinfo{year}{1972}),
  \bibinfo{pages}{288--295}.
\newblock


\bibitem[\protect\citeauthoryear{Muandet and Sch\"{o}lkopf}{Muandet and
  Sch\"{o}lkopf}{2013}]%
        {OCSMM2013}
\bibfield{author}{\bibinfo{person}{Krikamol Muandet} {and}
  \bibinfo{person}{Bernhard Sch\"{o}lkopf}.} \bibinfo{year}{2013}\natexlab{}.
\newblock \showarticletitle{One-class Support Measure Machines for Group
  Anomaly Detection}. In \bibinfo{booktitle}{\emph{Proceedings of the
  Twenty-Ninth Conference on Uncertainty in Artificial Intelligence}}.
  \bibinfo{pages}{449--458}.
\newblock


\bibitem[\protect\citeauthoryear{Omohundro}{Omohundro}{1989}]%
        {omohundro1989five}
\bibfield{author}{\bibinfo{person}{Stephen~M Omohundro}.}
  \bibinfo{year}{1989}\natexlab{}.
\newblock \bibinfo{booktitle}{\emph{Five balltree construction algorithms}}.
\newblock \bibinfo{publisher}{International Computer Science Institute
  Berkeley}.
\newblock


\bibitem[\protect\citeauthoryear{Pedregosa, Varoquaux, Gramfort, Michel,
  Thirion, Grisel, Blondel, Prettenhofer, Weiss, Dubourg, Vanderplas, Passos,
  Cournapeau, Brucher, Perrot, and Duchesnay}{Pedregosa et~al\mbox{.}}{2011}]%
        {scikit-learn}
\bibfield{author}{\bibinfo{person}{F. Pedregosa}, \bibinfo{person}{G.
  Varoquaux}, \bibinfo{person}{A. Gramfort}, \bibinfo{person}{V. Michel},
  \bibinfo{person}{B. Thirion}, \bibinfo{person}{O. Grisel},
  \bibinfo{person}{M. Blondel}, \bibinfo{person}{P. Prettenhofer},
  \bibinfo{person}{R. Weiss}, \bibinfo{person}{V. Dubourg}, \bibinfo{person}{J.
  Vanderplas}, \bibinfo{person}{A. Passos}, \bibinfo{person}{D. Cournapeau},
  \bibinfo{person}{M. Brucher}, \bibinfo{person}{M. Perrot}, {and}
  \bibinfo{person}{E. Duchesnay}.} \bibinfo{year}{2011}\natexlab{}.
\newblock \showarticletitle{Scikit-learn: Machine Learning in {P}ython}.
\newblock \bibinfo{journal}{\emph{Journal of Machine Learning Research}}
  \bibinfo{volume}{12} (\bibinfo{year}{2011}), \bibinfo{pages}{2825--2830}.
\newblock


\bibitem[\protect\citeauthoryear{Pestov}{Pestov}{2000}]%
        {Pestov}
\bibfield{author}{\bibinfo{person}{Vladimir Pestov}.}
  \bibinfo{year}{2000}\natexlab{}.
\newblock \showarticletitle{On the geometry of similarity search:
  Dimensionality curse and concentration of measure}.
\newblock \bibinfo{journal}{\emph{Inform. Process. Lett.}}
  \bibinfo{volume}{73}, \bibinfo{number}{1} (\bibinfo{year}{2000}),
  \bibinfo{pages}{47--51}.
\newblock


\bibitem[\protect\citeauthoryear{Qin, Ting, Zhu, and Lee}{Qin
  et~al\mbox{.}}{2019}]%
        {IsolationKernel-AAAI2019}
\bibfield{author}{\bibinfo{person}{Xiaoyu Qin}, \bibinfo{person}{Kai~Ming
  Ting}, \bibinfo{person}{Ye Zhu}, {and} \bibinfo{person}{Vincent Cheng~Siong
  Lee}.} \bibinfo{year}{2019}\natexlab{}.
\newblock \showarticletitle{Nearest-Neighbour-Induced Isolation Similarity and
  Its Impact on Density-Based Clustering}. In
  \bibinfo{booktitle}{\emph{Proceedings of The Thirty-Third AAAI Conference on
  Artificial Intelligence}}. \bibinfo{pages}{4755--4762}.
\newblock


\bibitem[\protect\citeauthoryear{Radovanovi\'{c}, Nanopoulos, and
  Ivanovi\'{c}}{Radovanovi\'{c} et~al\mbox{.}}{2010}]%
        {HubsInSpace-JMLR-2010}
\bibfield{author}{\bibinfo{person}{Milo{\v{s}} Radovanovi\'{c}},
  \bibinfo{person}{Alexandros Nanopoulos}, {and} \bibinfo{person}{Mirjana
  Ivanovi\'{c}}.} \bibinfo{year}{2010}\natexlab{}.
\newblock \showarticletitle{Hubs in Space: Popular Nearest Neighbors in
  High-Dimensional Data}.
\newblock \bibinfo{journal}{\emph{Journal of Machine Learning Research}}
  \bibinfo{volume}{11}, \bibinfo{number}{86} (\bibinfo{year}{2010}),
  \bibinfo{pages}{2487--2531}.
\newblock


\bibitem[\protect\citeauthoryear{Rahimi and Recht}{Rahimi and Recht}{2007}]%
        {RandomFeatures2007}
\bibfield{author}{\bibinfo{person}{Ali Rahimi} {and} \bibinfo{person}{Benjamin
  Recht}.} \bibinfo{year}{2007}\natexlab{}.
\newblock \showarticletitle{Random Features for Large-scale Kernel Machines}.
  In \bibinfo{booktitle}{\emph{Advances in Neural Information Processing
  Systems}}. \bibinfo{pages}{1177--1184}.
\newblock


\bibitem[\protect\citeauthoryear{Rodriguez and Laio}{Rodriguez and
  Laio}{2014}]%
        {DP2014clustering}
\bibfield{author}{\bibinfo{person}{Alex Rodriguez} {and}
  \bibinfo{person}{Alessandro Laio}.} \bibinfo{year}{2014}\natexlab{}.
\newblock \showarticletitle{Clustering by fast search and find of density
  peaks}.
\newblock \bibinfo{journal}{\emph{Science}} \bibinfo{volume}{344},
  \bibinfo{number}{6191} (\bibinfo{year}{2014}), \bibinfo{pages}{1492--1496}.
\newblock


\bibitem[\protect\citeauthoryear{Sch\"{o}lkopf, Platt, Shawe-Taylor, Smola, and
  Williamson}{Sch\"{o}lkopf et~al\mbox{.}}{2001}]%
        {OCSVM2001}
\bibfield{author}{\bibinfo{person}{Bernhard Sch\"{o}lkopf},
  \bibinfo{person}{John~C. Platt}, \bibinfo{person}{John~C. Shawe-Taylor},
  \bibinfo{person}{Alex~J. Smola}, {and} \bibinfo{person}{Robert~C.
  Williamson}.} \bibinfo{year}{2001}\natexlab{}.
\newblock \showarticletitle{Estimating the Support of a High-Dimensional
  Distribution}.
\newblock \bibinfo{journal}{\emph{Neural Computing}} \bibinfo{volume}{13},
  \bibinfo{number}{7} (\bibinfo{year}{2001}), \bibinfo{pages}{1443--1471}.
\newblock


\bibitem[\protect\citeauthoryear{Shaft and Ramakrishnan}{Shaft and
  Ramakrishnan}{2006}]%
        {NN-indexibility-TODS2006}
\bibfield{author}{\bibinfo{person}{Uri Shaft} {and} \bibinfo{person}{Raghu
  Ramakrishnan}.} \bibinfo{year}{2006}\natexlab{}.
\newblock \showarticletitle{Theory of Nearest Neighbors Indexability}.
\newblock \bibinfo{journal}{\emph{ACM Transactions on Database Systems}}
  \bibinfo{volume}{31}, \bibinfo{number}{3} (\bibinfo{year}{2006}),
  \bibinfo{pages}{814--838}.
\newblock


\bibitem[\protect\citeauthoryear{Talagrand}{Talagrand}{1996}]%
        {Talagrand}
\bibfield{author}{\bibinfo{person}{Michel Talagrand}.}
  \bibinfo{year}{1996}\natexlab{}.
\newblock \showarticletitle{A New Look at Independence}.
\newblock \bibinfo{journal}{\emph{The Annals of Probability}}
  \bibinfo{volume}{24}, \bibinfo{number}{1} (\bibinfo{year}{1996}),
  \bibinfo{pages}{1--34}.
\newblock


\bibitem[\protect\citeauthoryear{Ting, Wells, and Washio}{Ting
  et~al\mbox{.}}{2021}]%
        {IK-XFactor-2019}
\bibfield{author}{\bibinfo{person}{Kai~Ming Ting}, \bibinfo{person}{Jonathan~R.
  Wells}, {and} \bibinfo{person}{Takashi Washio}.}
  \bibinfo{year}{2021}\natexlab{}.
\newblock \showarticletitle{Isolation Kernel: The {X} Factor in Efficient and
  Effective Large Scale Online Kernel Learning}.
\newblock \bibinfo{journal}{\emph{Data Mining and Knowledge Discovery,
  https://doi.org/10.1007/s10618-021-00785-1}} (\bibinfo{year}{2021}).
\newblock


\bibitem[\protect\citeauthoryear{Ting, Xu, Washio, and Zhou}{Ting
  et~al\mbox{.}}{2020a}]%
        {IDK-KDD2020}
\bibfield{author}{\bibinfo{person}{Kai~Ming Ting}, \bibinfo{person}{Bi-Cun Xu},
  \bibinfo{person}{Takashi Washio}, {and} \bibinfo{person}{Zhi-Hua Zhou}.}
  \bibinfo{year}{2020}\natexlab{a}.
\newblock \showarticletitle{Isolation Distributional Kernel: A New Tool for
  Kernel Based Anomaly Detection}. In \bibinfo{booktitle}{\emph{Proceedings of
  the 26th ACM SIGKDD International Conference on Knowledge Discovery and Data
  Mining}}. \bibinfo{pages}{198--206}.
\newblock


\bibitem[\protect\citeauthoryear{Ting, Xu, Washio, and Zhou}{Ting
  et~al\mbox{.}}{2020b}]%
        {ting2020-IDK-GroupAnomalyDetection}
\bibfield{author}{\bibinfo{person}{Kai~Ming Ting}, \bibinfo{person}{Bi-Cun Xu},
  \bibinfo{person}{Takashi Washio}, {and} \bibinfo{person}{Zhi-Hua Zhou}.}
  \bibinfo{year}{2020}\natexlab{b}.
\newblock \showarticletitle{Isolation Distributional Kernel: A New Tool for
  Point and Group Anomaly Detection}.
\newblock \bibinfo{journal}{\emph{CoRR}} (\bibinfo{year}{2020}).
\newblock
\urldef\tempurl%
\url{https://arxiv.org/abs/2009.12196}
\showURL{%
\tempurl}


\bibitem[\protect\citeauthoryear{Ting, Zhu, and Zhou}{Ting
  et~al\mbox{.}}{2018}]%
        {ting2018IsolationKernel}
\bibfield{author}{\bibinfo{person}{Kai~Ming Ting}, \bibinfo{person}{Yue Zhu},
  {and} \bibinfo{person}{Zhi-Hua Zhou}.} \bibinfo{year}{2018}\natexlab{}.
\newblock \showarticletitle{Isolation Kernel and its effect on {SVM}}. In
  \bibinfo{booktitle}{\emph{Proceedings of the 24th ACM SIGKDD International
  Conference on Knowledge Discovery and Data Mining}}. ACM,
  \bibinfo{pages}{2329--2337}.
\newblock


\bibitem[\protect\citeauthoryear{van~der Maaten and Hinton}{van~der Maaten and
  Hinton}{2008}]%
        {Hinton2008Visualizing}
\bibfield{author}{\bibinfo{person}{Laurens van~der Maaten} {and}
  \bibinfo{person}{Geoffrey~E. Hinton}.} \bibinfo{year}{2008}\natexlab{}.
\newblock \showarticletitle{Visualizing High-Dimensional Data Using t-{SNE}}.
\newblock \bibinfo{journal}{\emph{Journal of Machine Learning Research}}
  \bibinfo{volume}{9}, \bibinfo{number}{2} (\bibinfo{year}{2008}),
  \bibinfo{pages}{2579--2605}.
\newblock


\bibitem[\protect\citeauthoryear{Vinh, Epps, and Bailey}{Vinh
  et~al\mbox{.}}{2010}]%
        {vinh2010information}
\bibfield{author}{\bibinfo{person}{Nguyen~Xuan Vinh}, \bibinfo{person}{Julien
  Epps}, {and} \bibinfo{person}{James Bailey}.}
  \bibinfo{year}{2010}\natexlab{}.
\newblock \showarticletitle{Information theoretic measures for clusterings
  comparison: Variants, properties, normalization and correction for chance}.
\newblock \bibinfo{journal}{\emph{The Journal of Machine Learning Research}}
  \bibinfo{volume}{11} (\bibinfo{year}{2010}), \bibinfo{pages}{2837--2854}.
\newblock


\bibitem[\protect\citeauthoryear{Wickelmaier}{Wickelmaier}{2003}]%
        {MDS-wickelmaier2003introduction}
\bibfield{author}{\bibinfo{person}{Florian Wickelmaier}.}
  \bibinfo{year}{2003}\natexlab{}.
\newblock \showarticletitle{An introduction to {MDS}}.
\newblock \bibinfo{journal}{\emph{Sound Quality Research Unit, Aalborg
  University}} (\bibinfo{year}{2003}).
\newblock


\bibitem[\protect\citeauthoryear{Williams and Seeger}{Williams and
  Seeger}{2001}]%
        {Nystrom_NIPS2000}
\bibfield{author}{\bibinfo{person}{Christopher K.~I. Williams} {and}
  \bibinfo{person}{Matthias Seeger}.} \bibinfo{year}{2001}\natexlab{}.
\newblock \showarticletitle{Using the Nystr\"{o}m Method to Speed Up Kernel
  Machines}.
\newblock In \bibinfo{booktitle}{\emph{Advances in Neural Information
  Processing Systems 13}}, \bibfield{editor}{\bibinfo{person}{T.~K. Leen},
  \bibinfo{person}{T.~G. Dietterich}, {and} \bibinfo{person}{V.~Tresp}} (Eds.).
  \bibinfo{publisher}{MIT Press}, \bibinfo{pages}{682--688}.
\newblock


\bibitem[\protect\citeauthoryear{Xu, Ting, and Zhou}{Xu et~al\mbox{.}}{2019}]%
        {IsolationSetKernel}
\bibfield{author}{\bibinfo{person}{Bi-Cun Xu}, \bibinfo{person}{Kai~Ming Ting},
  {and} \bibinfo{person}{Zhi-Hua Zhou}.} \bibinfo{year}{2019}\natexlab{}.
\newblock \showarticletitle{Isolation Set-Kernel and its application to
  Multi-Instance Learning}. In \bibinfo{booktitle}{\emph{Proceedings of the
  25th ACM SIGKDD International Conference on Knowledge Discovery and Data
  Mining}}. \bibinfo{pages}{941--949}.
\newblock


\bibitem[\protect\citeauthoryear{Zelnik-Manor and Perona}{Zelnik-Manor and
  Perona}{2005}]%
        {zelnik2005self}
\bibfield{author}{\bibinfo{person}{Lihi Zelnik-Manor} {and}
  \bibinfo{person}{Pietro Perona}.} \bibinfo{year}{2005}\natexlab{}.
\newblock \showarticletitle{Self-tuning spectral clustering}. In
  \bibinfo{booktitle}{\emph{Advances in neural information processing
  systems}}. \bibinfo{pages}{1601--1608}.
\newblock


\bibitem[\protect\citeauthoryear{Zhu and Ting}{Zhu and Ting}{2021}]%
        {IK-SNE-2021}
\bibfield{author}{\bibinfo{person}{Ye Zhu} {and} \bibinfo{person}{Kai~Ming
  Ting}.} \bibinfo{year}{2021}\natexlab{}.
\newblock \showarticletitle{Improving the Effectiveness and Efficiency of
  Stochastic Neighbour Embedding with Isolation Kernel}.
\newblock \bibinfo{journal}{\emph{Journal of Artificial Intelligence Research}}
   \bibinfo{volume}{71} (\bibinfo{year}{2021}), \bibinfo{pages}{667--695}.
\newblock


\end{thebibliography}

\end{document}